\documentclass{article}

\usepackage{microtype}
\usepackage{graphicx}
\usepackage{subcaption}
\usepackage{booktabs} 

\usepackage{hyperref}

\usepackage{bbding}
\usepackage{multirow}
\usepackage{amsfonts}

\usepackage[preprint]{icml2026}

\usepackage{amsmath}
\usepackage{amssymb}
\usepackage{mathtools}
\usepackage{amsthm}

\usepackage[capitalize,noabbrev]{cleveref}

\theoremstyle{plain}

\theoremstyle{definition}

\theoremstyle{remark}

\usepackage[textsize=tiny]{todonotes}

\usepackage{colortbl}
\usepackage{xcolor}
\definecolor{color_gray}{RGB}{229,229,229}
\definecolor{color_blue}{RGB}{252,182,165}
\definecolor{color_pink}{RGB}{255,217,178}
\definecolor{color_yellow}{RGB}{255,255,204}
\definecolor{color_blue1}{RGB}{135, 206, 235}
\cellcolor{color_blue} 
\cellcolor{color_pink} \cellcolor{color_yellow} 
\cellcolor{blue!8}
\icmltitlerunning{Preserving Geometric Fidelity in Efficient Transformer-Based PDE Solvers}

\begin{document}

\twocolumn[
  \icmltitle{Preserving Geometric Fidelity in Efficient Transformer-Based PDE Solvers}



  \icmlsetsymbol{equal}{*}

  \begin{icmlauthorlist}
    \icmlauthor{Zhuo Zhang}{yyy}
    \icmlauthor{Xi Yang}{yyy}
    \icmlauthor{Ying Miao}{pek}
    \icmlauthor{Xiaobin Hu}{nus}
    \icmlauthor{Yifu Gao}{yyy}
    \\
    \icmlauthor{Yong Yang}{tg}
    \icmlauthor{Canqun Yang}{comp}
    \icmlauthor{Boocheong Khoo}{nus}
  \end{icmlauthorlist}

  \icmlaffiliation{yyy}{National University of Defense Technology}
  
  \icmlaffiliation{comp}{National SuperComputer Center in Tianjin}

  \icmlaffiliation{nus}{National University of Singapore}
  \icmlaffiliation{tg}{Tiangong University}
  \icmlaffiliation{pek}{Peking University}

  \icmlcorrespondingauthor{Xi Yang}{yangxi1016@nudt.edu.cn}
  \icmlcorrespondingauthor{Xiaobin Hu}{ben0xiaobin0hu1@nus.edu.sg}

  \icmlkeywords{Machine Learning, ICML}

  \vskip 0.3in
]



\printAffiliationsAndNotice{}  

\begin{abstract}
Efficient Transformers for PDE solving typically aggregate mesh points into compact latent tokens to achieve linear complexity. We show that this aggregation acts as a spatial low-pass filter, systematically erasing high-frequency boundary details---a phenomenon we term \textit{geometric aliasing}. Through diagnostic experiments on standard benchmarks, we confirm that prediction errors of existing methods concentrate near complex geometric boundaries, validating this analysis. To address this, we build upon the slice-based attention framework and introduce two targeted modules. First, Spectrum-Preserving Geometric Attention injects multi-scale geometric encodings into both the slicing assignment and feature reconstruction stages, recovering boundary information while preserving $O(N)$ complexity. Second, a Taylor-Decomposed Feed-Forward Network routes features through linear and non-linear expert paths based on local geometric context, adapting computational capacity to the smoothness of the physical field. Experiments on four standard benchmarks and three large-scale industrial simulations 
(including a $\sim$300K-node 3D mesh) demonstrate consistent improvements, 
with up to \textbf{12.7\%} error reduction on standard tasks and \textbf{81.3\%} 
improvement on surface field prediction, with promising applications in industrial 
aerodynamic design and digital twin simulation.
\end{abstract}

\section{Introduction}

Partial Differential Equations (PDEs), derived from fundamental principles such as conservation of mass and energy, are ubiquitous in modeling complex phenomena across science and engineering, including fluid dynamics, structural mechanics, and climate modeling~\cite{cao2025solving}. Traditional numerical methods, such as the Finite Element Method (FEM)~\cite{femzienkiewicz2005finite} and Finite Volume Method (FVM)~\cite{fvmeymard2000finite}, approximate solutions by discretizing PDEs onto meshes~\cite{nudt1wanggexinhaichen2025gnnrl}. However, their reliance on dense mesh discretization imposes a heavy computational burden, often requiring much time and effort for multi-physics problems necessitating fine spatiotemporal resolution~\cite{mm410.1145/3746027.3755436,mm510.1145/3746027.3758161}. Recently, neural PDE solvers have emerged as a promising paradigm, capable of learning resolution-invariant solution mappings between infinite-dimensional function spaces~\cite{liangzinoicmlDBLP:conf/icml/WangX0Y25,fuliyesuanziicmlDBLP:conf/icml/LiY025}. By training on offline data, these operators can predict families of solutions under varying parameters orders of magnitude faster than traditional solvers, offering a promising direction for real-time scientific computing.

\begin{figure}[htbp]
	\centering
	\includegraphics[width=0.95\linewidth]{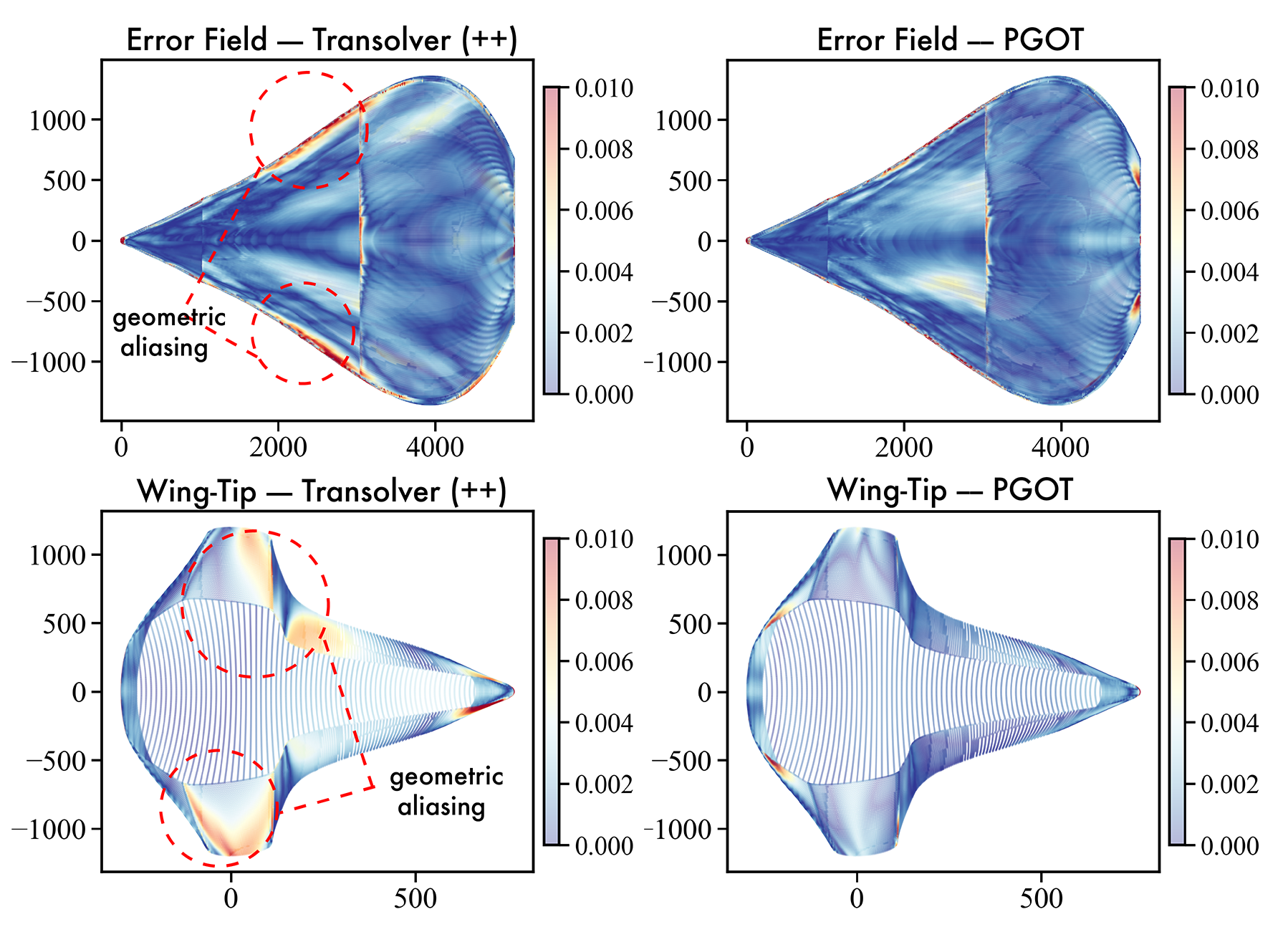}
	\caption{Qualitative analysis of geometric aliasing on the 3D AirCraft wing. (Top) Global error field comparison. (Bottom) Detailed view of the high-curvature wing-tip region.}
	\label{hundie}
\end{figure}

\begin{figure}[htbp]
	\centering
	\includegraphics[width=1\linewidth]{leidatu_spectrum.png}
	\caption{(a) Geometric aliasing diagnostic on AirCraft. (a.1) Transolver exhibits higher error power in the high-frequency band (shaded); (a.2) PGOT improvement increases with curvature complexity. (b) Performance comparison across benchmarks.}
	\label{qipao}
\end{figure}
Early spectral-based neural operators, such as the Fourier Neural Operator (FNO)~\cite{fnoDBLP:conf/iclr/LiKALBSA21}, rely on the fast fourier transform. While they possess the remarkable ability to learn resolution-independent operators, they are inherently limited to regular grids, restricting their applicability to complex geometries and unstructured meshes. To address this, graph-based neural operators~\cite{GNOTDBLP:conf/icml/HaoWSYDLCSZ23, GINODBLP:conf/nips/LiKCLKONS0AA23} treat input and output functions as graphs, employing message-passing mechanisms to generalize across different domain geometries. However, graph methods are often constrained by local receptive fields, making it difficult to effectively capture global long-range dependencies and suffer from low efficiency on large-scale nodes.

This necessitates the adoption of Transformers~\cite{vaswani2017attention}. The self-attention mechanism in Transformers is mathematically analogous to learning a global Green’s function~\cite{galerkinDBLP:conf/nips/Cao21}, making it ideally suited for modeling non-local physical interactions~\cite{suanzinipsDBLP:conf/nips/AlkinFSGHB24,fuliyenipsDBLP:conf/nips/KoshizukaFTS24}. However, a fundamental scalability barrier remains: the standard self-attention mechanism exhibits quadratic computational complexity $\mathcal{O}(N^2)$ with respect to the number of mesh points $N$. Given that scientific simulations typically involve $10^5$ to $10^7$ grid points, standard Transformers are computationally intractable for high-resolution physics. Consequently, the research frontier has shifted towards efficient linear transformers~\cite{oformerDBLP:journals/tmlr/LiMF23,trans++DBLP:conf/icml/LuoWZXD0L25,transolverDBLP:conf/icml/WuLW0L24}. Recent state-of-the-art methods employ strategies such as low-rank approximations or token clustering to reduce the effective sequence length, achieving linear complexity $\mathcal{O}(N)$~\cite{GFNETDBLP:conf/nips/RaoZZLZ21,ONODBLP:conf/icml/XiaoHLD024}. Despite significant progress, existing efficient Transformers share a fundamental blind spot in that none explicitly addresses the spectral consequence of token aggregation, treating geometry merely as an auxiliary input rather than a first-class signal throughout the network. While these architectures successfully improve throughput, they often do so at the cost of geometric and physical fidelity, as follows:

\textbf{\textit{L1):}} Aggregating local mesh points into latent tokens implicitly acts as a spatial low-pass filter, causing geometric aliasing where high-frequency boundary details are erased.
    
\textbf{\textit{L2):}} Allocating a uniform computational budget across the domain neglects physical heterogeneity, resulting in over-smoothed shocks in discontinuity regions and redundant computations in smooth flows.

To bridge the gap between computational efficiency and 
high-fidelity physical modeling, we present the Physics-Geometry 
Operator Transformer (PGOT). Unlike prior methods that treat 
geometry merely as a supplementary input, PGOT intertwines 
physical state aggregation with explicit geometric reconstruction.  As illustrated in Figure~\ref{hundie}, PGOT effectively suppresses 
geometric aliasing artifacts in high-curvature regions such as 
wing tips and leading edges. By injecting multi-scale geometric 
embeddings during reconstruction, PGOT eliminates aliasing while 
maintaining $\mathcal{O}(N)$ complexity \textbf{\textit{(Solving L1)}}, 
and dynamically routes computational paths based on local flow 
complexity to handle physical heterogeneity 
\textbf{\textit{(Solving L2)}}. As illustrated in 
Figure~\ref{qipao}, panel (a) provides diagnostic evidence for 
geometric aliasing and panel (b) summarizes the overall 
performance comparison.

Our main contributions are as follows:

\begin{itemize}
    \item Our proposed PGOT framework addresses the geometric aliasing bottleneck inherent in complex mesh modeling, thereby reconciling computational efficiency with physical fidelity.
    
    \item We propose the SpecGeo-Attention mechanism to decouple global physical aggregation from local geometric reconstruction via a ``physics slicing-geometry injection" paradigm, thereby preserving complex boundary features with high fidelity.
    
    \item The Taylor-decomposed Feed-Forward Network adaptively routes computations to linear or non-linear paths based on flow heterogeneity, capturing multi-scale physical features with enhanced interpretability.
\end{itemize}

\section{Related Work}

\subsection{Physics-Informed Neural Networks}
Physics-Informed Neural Networks (PINNs)~\cite{pinnraissi2019physics} represent a foundational paradigm for solving PDEs by embedding governing equations directly into the loss function. PINNs parameterize the solution as a neural network and minimize the PDE residual at collocation points, enabling mesh-free training without labeled simulation data. However, PINNs face fundamental limitations for the problems we target~\cite{mm110.1145/3746027.3755312,mm210.1145/3746027.3755829,mm310.1145/3746027.3754569}. First, PINNs solve individual PDE instances, requiring complete retraining for each new parameter configuration or boundary condition—a prohibitive cost for design optimization requiring thousands of evaluations. Second, for complex geometries and multi-physics coupling, PINNs frequently encounter ill-conditioning issues~\cite{me2zhang2025pseudo}, often leading to convergence failures or accuracy degradation. Although recent research has alleviated these problems through parameterization~\cite{canshuhuaicmlDBLP:conf/icml/ChoJLL00P24}, meta-learning~\cite{yuanxuexicheng2025meta}, and transfer learning~\cite{wyztransferng2025transfer} strategies, PINNs still struggle to maintain accuracy and efficiency when the configuration space becomes sufficiently large. These limitations motivate the development of the neural operator paradigm.

\subsection{Neural Operators for PDEs}

Neural operators have emerged as a powerful paradigm for learning resolution-invariant mappings between infinite-dimensional function spaces~\citep{suanzidiyijukovachki2023neural}. DeepONet~\citep{deeponetlu2021learning} pioneered this field by approximating operators through a Branch-Trunk architecture. Subsequently, the Fourier Neural Operator (FNO)~\citep{fnoDBLP:conf/iclr/LiKALBSA21} established new benchmarks by parameterizing the integral kernel directly in the frequency domain using FFT. To address the limitations of standard FNO, numerous variants have been proposed. U-NO~\citep{unoDBLP:journals/tmlr/RahmanRA23} and U-FNO~\citep{unfnowen2022u} integrate U-Net structures to capture multi-scale features, while F-FNO~\citep{ffnoDBLP:conf/iclr/TranMXO23} improves computational efficiency by factorizing the Fourier representation across dimensions. To handle irregular geometries beyond rectangular grids, Geo-FNO~\citep{geofnoli2023fourier} maps physical domains to latent uniform grids, while LSM~\citep{LSMDBLP:conf/icml/WuHLWL23} and LNO~\citep{LNODBLP:conf/nips/WangW24} apply spectral methods within learned low-dimensional latent spaces to tackle high-dimensional PDE problems. Furthermore, to mitigate the global nature of Fourier transforms that typically leads to loss of local details, wavelet-based methods such as WMT~\citep{MWTDBLP:conf/nips/GuptaXB21} and WNO~\citep{WNOtripura2023wavelet} have been introduced to leverage the time-frequency localization properties of wavelets for capturing local variations. For modeling complex unstructured meshes, geometric deep learning offers an alternative perspective. Graph Neural Operators (GNO)~\citep{GNOli2020neural} and MeshGraphNet~\citep{MeshGraphDBLP:conf/iclr/PfaffFSB21} treat physical domains as graphs, employing message-passing mechanisms to handle complex boundaries. To combine global spectral processing capabilities with local geometric flexibility, hybrid architectures have emerged. For instance, GINO~\citep{GINODBLP:conf/nips/LiKCLKONS0AA23} integrates Geo-FNO's global spectral capabilities with GNO's local graph processing, while 3D-GeoCA~\citep{3dgeoDBLP:conf/ijcai/DengLXHM24} further incorporates pre-trained 3D vision backbones to enhance geometric feature extraction. However, graph-based methods typically suffer from geometric instability~\citep{gnnbuhaomorris2023geometric, trans++DBLP:conf/icml/LuoWZXD0L25,wen2025gaot} and incur high computational overhead when scaling to million-scale mesh points due to neighborhood search and message passing.
\subsection{Transformers for PDEs}
The Transformer architecture has garnered significant attention in scientific computing due to its global receptive field and superior modeling capabilities. Early works such as OFormer~\citep{oformerDBLP:journals/tmlr/LiMF23} and HT-Net~\citep{htnetliu2024mitigating} demonstrated the potential of attention mechanisms in capturing non-local physical interactions. However, the quadratic complexity $\mathcal{O}(N^2)$ of standard attention remains a bottleneck restricting its scalability. To achieve linear complexity $\mathcal{O}(N)$, recent research has explored diverse pathways. AFNO~\citep{AFNODBLP:conf/iclr/GuibasMLTAC22}, FNet ~\citep{FNETDBLP:conf/naacl/Lee-ThorpAEO22}, and GFNet \cite{GFNETDBLP:conf/nips/RaoZZLZ21} attempt token mixing in the Fourier domain, while SAOT~\citep{SAOTzhou2025dual} further integrates wavelet attention to mitigate aliasing issues; simultaneously, Galerkin Transformer~\citep{galerkinDBLP:conf/nips/Cao21}, GNOT~\citep{GNOTDBLP:conf/icml/HaoWSYDLCSZ23}, and ONO~\citep{ONODBLP:conf/icml/XiaoHLD024} utilize kernel tricks or low-rank approximations to achieve linearization; whereas methods represented by FactFormer~\citep{FactFormerDBLP:conf/nips/LiSF23}, 
Transolver~\citep{trans++DBLP:conf/icml/LuoWZXD0L25,transolverDBLP:conf/icml/WuLW0L24}, and 
GAOT~\citep{wen2025gaot} employ axial decomposition or token aggregation strategies, respectively. Despite their significant efficiency improvements, these dimensionality reduction or aggregation operations act as implicit spatial low-pass filters, inevitably discarding high-frequency geometric details. Furthermore, they typically allocate a uniform computational budget across the entire domain, failing to adapt to physical heterogeneity in flow fields.

\section{Method}
\label{sec:method}

\begin{figure}[htbp]
	\centering
	\includegraphics[width=1\linewidth]{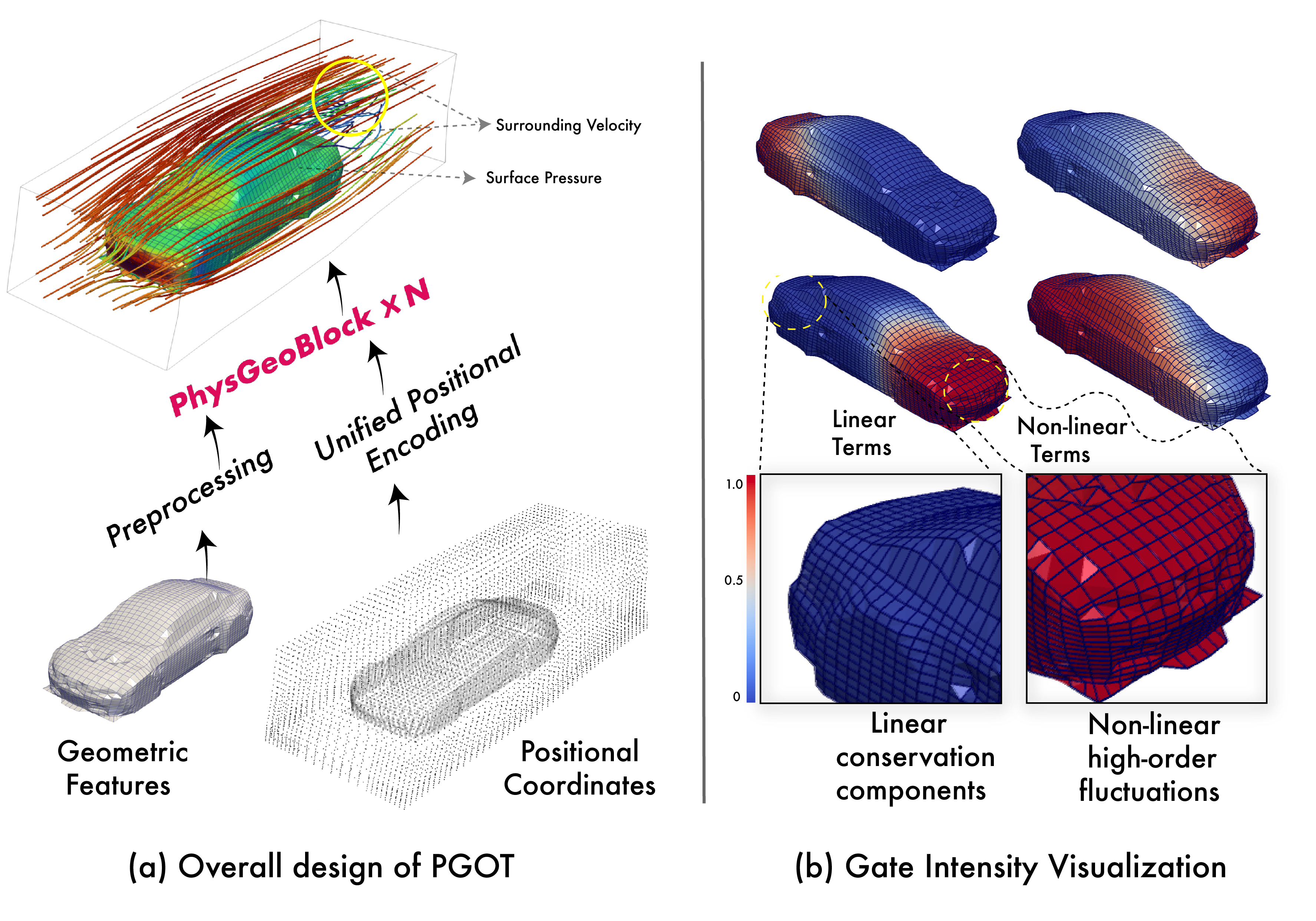}
	\caption{Overall architecture of PGOT. (a) The framework explicitly integrates multi-scale geometry via stacked PhysGeoBlocks to reconstruct velocity and pressure fields on complex 3D meshes. (b) Visualization of TaylorDecomp-FFN. The Linear Expert (blue) captures smooth conservation dynamics, while the Non-linear Expert (red) targets high-order fluctuations.}

	\label{fig:main_arch}
\end{figure}

In this section, we present the design of the Physics-Geometry Operator Transformer (PGOT). We first analyze the geometric aliasing bottleneck inherent in existing efficient architectures. Subsequently, we detail the two core components: Spectrum-Preserving Geometric Attention (SpecGeo-Attention) and the Taylor-Decomposed Feed-Forward Network (TaylorDecomp-FFN). The overall architecture is illustrated in \figurename{}~\ref{fig:main_arch}.

\begin{figure*}[htbp]
	\centering
	\includegraphics[width=1\linewidth]{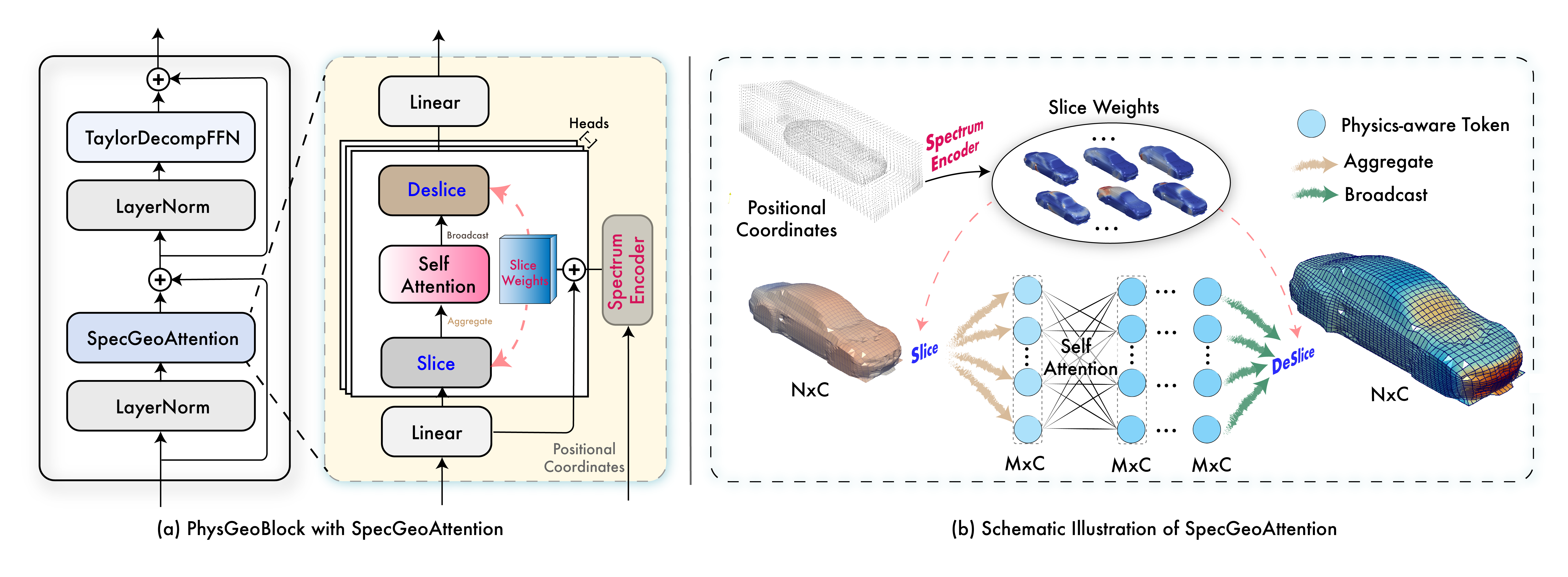}
\caption{Architecture of PhysGeoBlock and SpecGeo-Attention. (a) The PhysGeoBlock integrates explicit geometric coordinates into the SpecGeo-Attention and TaylorDecomp-FFN layers. (b) The ``physics slicing-geometry injection" paradigm. A Spectrum Encoder generates geometry-aware weights to aggregate $N$ mesh points into $M$ latent tokens (Slice) and reconstruct them (DeSlice). This design achieves linear complexity $\mathcal{O}(N)$ while preserving multi-scale geometric fidelity.}

	\label{physGeoBlock}
\end{figure*}
\subsection{Problem Formulation}

We consider a physical system governed by partial differential equations (PDEs), where the solution operator $\mathcal{G}^\dagger$ maps an input function space to an output function space: $\mathcal{G}^\dagger: \mathcal{A}(\Omega; \mathbb{R}^{d_a}) \rightarrow \mathcal{U}(\Omega; \mathbb{R}^{d_u})$. In practice, the continuous domain $\Omega \subset \mathbb{R}^{d}$ is discretized into $N$ mesh points $\mathbf{G} = \{\mathbf{g}_i\}_{i=1}^{N}$. The input and output functions are sampled as $\mathbf{a} \in \mathbb{R}^{N \times d_a}$ and $\mathbf{u} \in \mathbb{R}^{N \times d_u}$, respectively. Our goal is to learn a parameterized neural operator $\mathcal{G}_\theta$ that approximates the ground truth operator $\mathcal{G}^\dagger$.

Existing efficient Transformers achieve linear complexity through feature aggregation. However, this operation acts as a low-pass filter in the frequency domain. Let $f(\mathbf{g})$ be the geometric signal. Aggregation can be modeled as a convolution with a smoothing kernel $K_\sigma$. According to the convolution theorem, the frequency response is $\hat{f}_{agg}(\omega) = \hat{K}_\sigma(\omega) \cdot \hat{f}(\omega)$. When the spatial frequency $|\omega|$ exceeds the cutoff frequency $\omega_c \propto 1/\sigma$, $\hat{K}_\sigma(\omega) \approx 0$, causing high-frequency geometric details to be irreversibly erased. We term this phenomenon geometric aliasing, identifying it as a fundamental bottleneck limiting performance on complex boundaries. PGOT addresses this via explicit multi-scale geometric reconstruction.

\subsection{PGOT}

Recent work has demonstrated that slice-based attention, which aggregates mesh points into learnable physics-aware tokens, can achieve linear complexity for PDE solving~\citep{transolverDBLP:conf/icml/WuLW0L24}. However, this aggregation implicitly acts as a spatial low-pass filter, causing geometric aliasing at complex boundaries. To address this limitation, PGOT introduces Spectrum-Preserving Geometric Attention and Taylor-Decomposed Feed-Forward Networks.
\begin{equation}
    \mathbf{X}^0 = \mathcal{P}_\theta\left(\mathbf{a} \oplus \mathcal{E}_{\text{pos}}(\mathbf{G})\right),
\end{equation}
where $\oplus$ denotes concatenation, and $\mathcal{P}_\theta$ is a lifting MLP. We employ a Unified Reference Point Encoding $\mathcal{E}_{\text{pos}}$ to ensure discretization invariance.

The processor consists of $L$ stacked \textit{PhysGeoBlocks}, as depicted in \figurename{}~\ref{physGeoBlock}(a). Unlike prior methods that use coordinates only at the input, PGOT explicitly injects geometry $\mathbf{G}$ at every layer:
\begin{align}
    \hat{\mathbf{X}}^l &= \mathbf{X}^{l-1} + \text{SpecGeo-Attn}\left(\text{LN}(\mathbf{X}^{l-1}); \mathbf{G}\right), \\
    \mathbf{X}^l &= \hat{\mathbf{X}}^l + \text{TaylorDecomp}\left(\text{LN}(\hat{\mathbf{X}}^l); \mathbf{G}\right),
\end{align}
where $\text{LN}(\cdot)$ denotes Layer Normalization. This design ensures that geometric guidance persists throughout the deep network evolution.

\subsubsection{Spectrum-Preserving Geometric Attention}

As illustrated in \figurename{}~\ref{physGeoBlock}(b), the core philosophy of SpecGeo-Attention is to decouple global physical state aggregation from local geometric reconstruction. 

\paragraph{Multi-Scale Geometric Spectral Encoding.} To reconstruct high-frequency information lost during aggregation, we inject multi-scale geometric encodings prior to slicing. We design a bank of encoders for $S$ scales:
\begin{equation}
    \mathbf{h}_s = \phi_s\left(10^{s-1} \cdot \mathbf{g}\right), \quad s = 1, \ldots, S,
\end{equation}
where $\phi_s$ is a two-layer MLP. The exponential scaling factor $10^{s-1}$ ensures coverage of diverse spatial frequencies: $s=1$ captures macro-structures, while higher scales capture micro-boundary details. These features are fused via concatenation to preserve full spectral information:
\begin{equation}
    \Phi_{\text{geo}}(\mathbf{g}) = \sigma_{\text{act}}\left(\mathbf{W}_{\text{fuse}} \cdot \left[\mathbf{h}_1; \mathbf{h}_2; \ldots; \mathbf{h}_S\right]\right),
\end{equation}
where $[\cdot;\cdot]$ denotes concatenation.

\paragraph{Geometry-Informed Physics Slicing.} The geometric encoding is injected into the physical features to guide the slicing process:
\begin{equation}
    \tilde{\mathbf{X}}_q = \mathbf{X} \mathbf{W}_x + \Phi_{\text{geo}}(\mathbf{G}).
\end{equation}
We then compute the soft assignment matrix $\mathbf{A} \in \mathbb{R}^{N \times M}$ mapping $N$ mesh points to $M$ latent physical states:
\begin{equation}
    \mathbf{A}_{ij} = \frac{\exp\left([\tilde{\mathbf{X}}_q]_i \cdot \mathbf{w}_j / \tau\right)}{\sum_{k=1}^{M} \exp\left([\tilde{\mathbf{X}}_q]_i \cdot \mathbf{w}_k / \tau\right)},
\end{equation}
where $\{\mathbf{w}_j\}$ are learnable slice prototypes. Since $\mathbf{A}$ is derived from geometry-enhanced features, the clustering respects complex boundaries (e.g., distinguishing the pressure side from the suction side of an airfoil).

Physical features are then aggregated into latent tokens $\mathbf{Z} \in \mathbb{R}^{M \times C}$ via $\mathbf{Z} = \mathbf{D}^{-1} \mathbf{A}^\top (\mathbf{X} \mathbf{W}_f)$, processed by standard Multi-Head Self-Attention (MHSA), and reconstructed back to the mesh domain:
\begin{equation}
    \mathbf{X}' = \mathbf{A} \cdot \text{MHSA}(\mathbf{Z}).
\end{equation}
This entire process maintains $\mathcal{O}(N)$ complexity while preserving geometric fidelity.

\subsubsection{TaylorDecomp-FFN}

Physical fields exhibit significant spatial heterogeneity: smooth laminar regions follow low-order linear laws, while shocks and discontinuities are governed by high-order non-linear dynamics. Drawing inspiration from Taylor series expansion, we approximate the solution operator at any spatial point $\mathbf{g}$ by decomposing it into low-order and high-order components:
\begin{equation}
    \mathbf{u}(\mathbf{g}) \approx \underbrace{\mathbf{u}(\mathbf{g}_0) + \nabla\mathbf{u}|_{\mathbf{g}_0}(\mathbf{g}-\mathbf{g}_0)}_{\text{Linear Terms}} + \underbrace{\mathcal{R}_{high}(\mathbf{g})}_{\text{Non-linear Residuals}}.
\end{equation}
To adaptively model these terms, we propose the TaylorDecomp-FFN, which formulates the feature transformation as a dynamic composition of two expert paths:
\begin{equation}
    \mathbf{X}_{out} = (\mathbf{1} - \boldsymbol{\alpha}(\mathbf{g})) \odot \mathcal{F}_{\text{lin}}(\mathbf{X}) + \boldsymbol{\alpha}(\mathbf{g}) \odot \mathcal{F}_{\text{non}}(\mathbf{X}).
\end{equation}
The architecture explicitly instantiates the theoretical decomposition above. The Linear Expert ($\mathcal{F}_{\text{lin}}$), corresponding to the low-order Taylor terms, consists of a streamlined linear projection with dropout to efficiently capture smooth conservation dynamics:
\begin{equation}
    \mathcal{F}_{\text{lin}}(\mathbf{x}) = \mathbf{W}_{l2}(\text{Dropout}(\mathbf{W}_{l1}\mathbf{x})).
\end{equation}
In parallel, the Non-linear Expert ($\mathcal{F}_{\text{non}}$) targets the high-order residuals, utilizing a deeper MLP with activation functions to resolve sharp gradients and discontinuities:
\begin{equation}
    \mathcal{F}_{\text{non}}(\mathbf{x}) = \mathbf{W}_{n2}(\sigma(\mathbf{W}_{n1}\mathbf{x})).
\end{equation}

The dynamic routing between these two paths is governed by a spatial gate $\boldsymbol{\alpha}(\mathbf{g}) \in [0, 1]^C$. To ensure invariance to geometric transformations such as translation and rotation, we encode each point's position relative to its assigned slice centroid rather than using absolute coordinates:
\begin{equation}
    \boldsymbol{\alpha}(\mathbf{g}) = \text{Sigmoid}\left(\text{MLP}_{gate}\left(\text{PosEmbed}(\mathbf{g} - \bar{\mathbf{g}}_{\text{slice}})\right)\right),
\end{equation}
where $\bar{\mathbf{g}}_{\text{slice}} = \sum_{i} \mathbf{A}_{ij} \mathbf{g}_i / \sum_{i} \mathbf{A}_{ij}$ is the soft centroid of the slice to which point $\mathbf{g}$ is most strongly assigned. This local coordinate embedding captures each point's relative displacement within its physical region, enabling the gate to detect boundary proximity and flow complexity from geometric context alone.

We employ \textit{unconstrained regime selection} where the gate $\boldsymbol{\alpha}$ operates independently per channel. This flexibility allows the model to fully activate the non-linear path in shock regions ($\alpha \to 1$), rely on the linear path in smooth flows ($\alpha \to 0$), or leverage a hybrid combination ($\alpha \approx 0.5$) in complex transition zones.

\section{Experiments}
\label{sec:experiments}

We evaluate PGOT on four standard benchmarks and three large-scale industrial simulations, covering diverse geometric settings from point clouds to complex 3D unstructured meshes with up to $\sim$300K nodes.

\subsection{Experimental Setup}

\paragraph{Datasets.} Our evaluation encompasses seven benchmarks with varying geometric complexities. The standard benchmarks~\cite{fnoDBLP:conf/iclr/LiKALBSA21, geofnoli2023fourier} include Elasticity (inner stress prediction on 2D point clouds), Plasticity (temporal displacement forecasting under external forces), Airfoil (transonic flow simulation with shock waves), and Pipe (velocity field estimation in curved structures). For industrial-scale validation, we employ Shape-Net Car~\cite{CARumetani2018learning} for automotive aerodynamic simulation on hybrid 3D geometries, and AirfRANS~\cite{AIRANSbonnet2022airfrans} for high-fidelity RANS simulation on NACA airfoils under varying Reynolds numbers and angles of attack. For large-scale 3D validation, we additionally evaluate on the AirCraft 
benchmark~\cite{trans++DBLP:conf/icml/LuoWZXD0L25}, a 3D wing surface 
prediction task featuring $\sim$300K unstructured mesh nodes per sample 
under varying Mach numbers and angles of attack. Detailed dataset descriptions are provided in Appendix~B.1.

\paragraph{Baselines.} We compare against representative methods from three categories: spectral neural operators (FNO~\cite{fnoDBLP:conf/iclr/LiKALBSA21}, Geo-FNO~\cite{geofnoli2023fourier}, U-NO~\cite{unoDBLP:journals/tmlr/RahmanRA23}, F-FNO~\cite{ffnoDBLP:conf/iclr/TranMXO23}, LSM~\cite{LSMDBLP:conf/icml/WuHLWL23}, WMT~\cite{MWTDBLP:conf/nips/GuptaXB21}); Transformer-based approaches (Galerkin Transformer~\cite{galerkinDBLP:conf/nips/Cao21}, OFormer~\cite{oformerDBLP:journals/tmlr/LiMF23}, GNOT~\cite{GNOTDBLP:conf/icml/HaoWSYDLCSZ23}, ONO~\cite{ONODBLP:conf/icml/XiaoHLD024}, FactFormer~\cite{FactFormerDBLP:conf/nips/LiSF23}, 
IPOT~\cite{ipot2024inducing}, Transolver (++)~\cite{transolverDBLP:conf/icml/WuLW0L24,trans++DBLP:conf/icml/LuoWZXD0L25}, SAOT~\cite{SAOTzhou2025dual}); and geometric deep learning methods (PointNet~\cite{qi2017pointnet}, Graph U-Net~\cite{graphunet2019graph}, GraphSAGE~\cite{sagehamilton2017inductive}, MeshGraphNet~\cite{MeshGraphDBLP:conf/iclr/PfaffFSB21}, GNO~\cite{GNOli2020neural}, GINO~\cite{GINODBLP:conf/nips/LiKCLKONS0AA23}, 3D-GeoCA~\cite{3dgeoDBLP:conf/ijcai/DengLXHM24}). Note that Transolver and Transolver++ share an identical backbone 
architecture, differing only in the slice-weight sampling strategy 
and a multi-GPU parallelism framework. Since our evaluation is 
conducted on a single GPU, we adopt the Transolver++ network 
architecture (without multi-GPU parallelism) and refer to it 
uniformly as Transolver (++) throughout this paper. See Appendix~B.4 for details.

\paragraph{Metrics and Implementation.} We adopt the relative $L^2$ error as the primary evaluation metric. For industrial benchmarks, we additionally report aerodynamic coefficient errors and Spearman's rank correlation. PGOT is configured with $L=8$ PhysGeoBlocks, with other hyperparameters adapted to each task. All experiments are conducted on NVIDIA A100 GPUs and repeated three times. Comprehensive details on metrics, training configurations, and hyperparameters are provided in Appendix~B.2 and Appendix~B.3. 
\subsection{Main Results}

\begin{table*}[htbp]
\centering
\caption{Performance comparison on standard benchmarks. The relative $L^2$ error is reported. The best results are highlighted in \textbf{bold}, and the second best are \underline{underlined}. \colorbox{gray!20}{Gray background} indicates our proposed method. Improvement denotes the reduction in relative error.}
\label{tab:standard}

\begin{tabular}{lcccc}
\toprule
\multirow{2}{*}{\textbf{Model}} & \multicolumn{3}{c}{\textbf{Structured Mesh}} & \textbf{Point Cloud} \\
\cmidrule(lr){2-4} \cmidrule(lr){5-5}
 & Plasticity & Pipe & Airfoil & Elasticity \\
\midrule
WMT~\citeyearpar{MWTDBLP:conf/nips/GuptaXB21} & 0.0076 & 0.0077 & 0.0075 & 0.0359 \\
U-FNO~\citeyearpar{unfnowen2022u}& 0.0039 & 0.0056 & 0.0269 & 0.0239 \\
Geo-FNO~\citeyearpar{geofnoli2023fourier} & 0.0074 & 0.0067 & 0.0138 & 0.0229 \\
U-NO~\citeyearpar{unoDBLP:journals/tmlr/RahmanRA23} & 0.0034 & 0.0100 & 0.0078 & 0.0258 \\
F-FNO~\citeyearpar{ffnoDBLP:conf/iclr/TranMXO23}& 0.0047 & 0.0070 & 0.0078 & 0.0263 \\
LSM~\citeyearpar{LSMDBLP:conf/icml/WuHLWL23} & 0.0025 & 0.0050 & 0.0059 & 0.0218 \\
\midrule
Galerkin~\citeyearpar{galerkinDBLP:conf/nips/Cao21} & 0.0120 & 0.0098 & 0.0118 & 0.0240 \\
HT-Net~\citeyearpar{htnetliu2024mitigating} & 0.0333 & 0.0059 & 0.0065 & / \\
OFormer~\citeyearpar{oformerDBLP:journals/tmlr/LiMF23} & 0.0017 & 0.0168 & 0.0183 & 0.0183 \\
GNOT~\citeyearpar{GNOTDBLP:conf/icml/HaoWSYDLCSZ23} & 0.0336 & 0.0047 & 0.0076 & 0.0086 \\
FactFormer~\citeyearpar{FactFormerDBLP:conf/nips/LiSF23} & 0.0312 & 0.0060 & 0.0071 & / \\
ONO~\citeyearpar{ONODBLP:conf/icml/XiaoHLD024} & 0.0048 & 0.0052 & 0.0061 & 0.0118 \\
IPOT~\citeyearpar{ipot2024inducing} & 0.0033 & / & 0.0088 & 0.0156 \\
Transolver (++)~\citeyearpar{transolverDBLP:conf/icml/WuLW0L24,trans++DBLP:conf/icml/LuoWZXD0L25} & \underline{0.0013} & \underline{0.0043} & 0.0053 & \underline{0.0079} \\
SAOT~\citeyearpar{SAOTzhou2025dual} & 0.0014 & 0.0062 & \underline{0.0051} & 0.0090 \\
\midrule
\cellcolor{gray!20}\textbf{PGOT (Ours)} & \cellcolor{gray!20}\textbf{0.0012} & \cellcolor{gray!20}\textbf{0.0039} & \cellcolor{gray!20}\textbf{0.0046} & \cellcolor{gray!20}\textbf{0.0069} \\
\midrule
\textbf{Improvement} & 7.7\% & 9.3\% & 9.8\% & 12.7\% \\
\bottomrule
\end{tabular}

\end{table*}

\begin{figure}[htbp]
    \centering
    \includegraphics[width=0.5\textwidth]{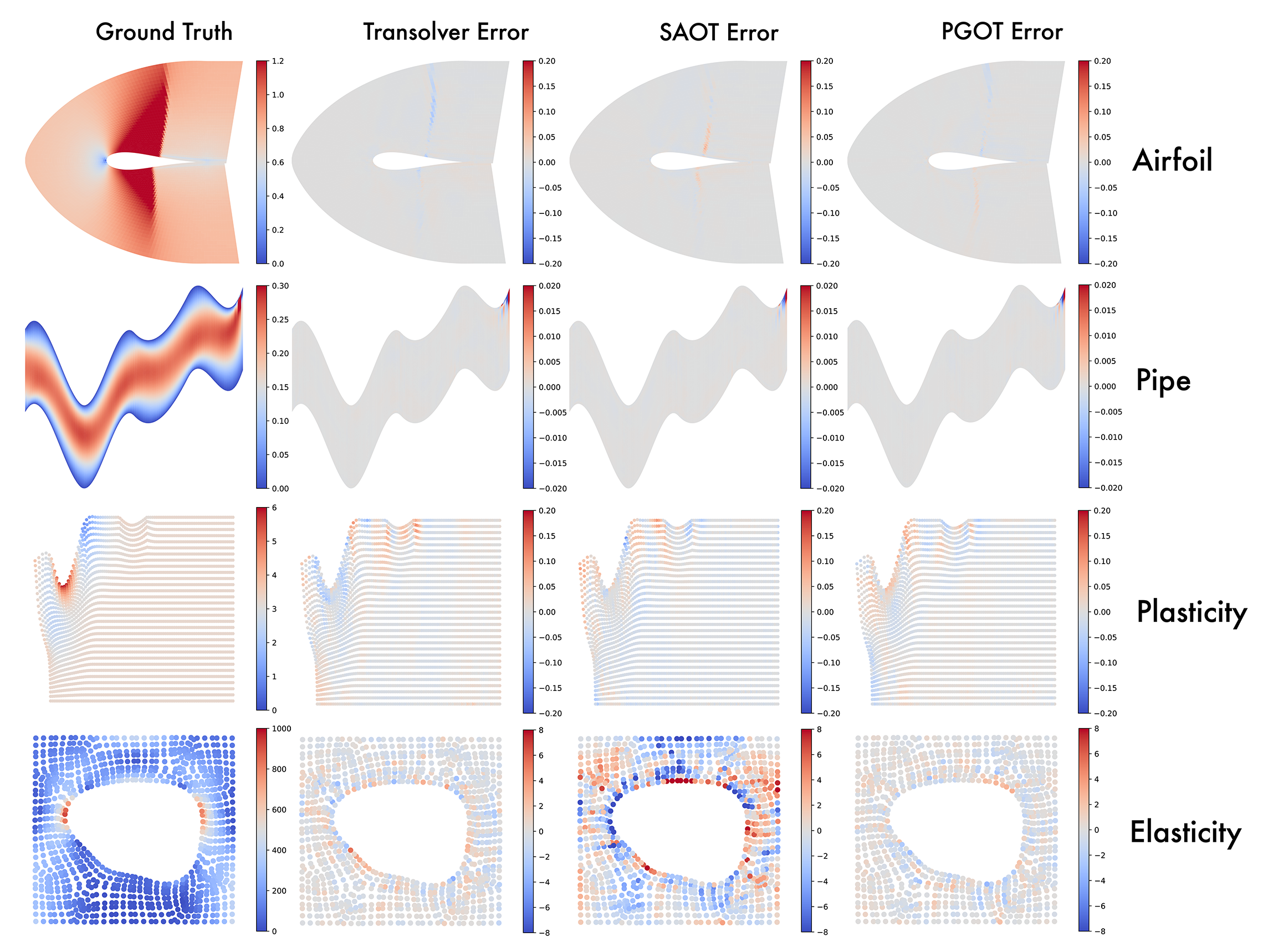}
    \caption{Visualization of prediction errors on standard benchmarks. The first column shows ground truth fields, and the remaining columns display prediction errors.}
    \label{fig:vis_standard}
    
\end{figure}

\begin{table*}[htbp]
\centering
\small
\caption{Performance comparison on practical design tasks. For Shape-Net Car, relative $L^2$ of surrounding (Volume) and surface (Surf) physics fields is recorded. For AirfRANS, MSE of Volume and Surf fields is recorded. For AirCraft , the last three decimal places are retained to facilitate direct comparison with published results. The relative $L^2$ of drag coefficient ($C_D$) and lift coefficient ($C_L$) is also recorded, along with their Spearman's rank correlations $\rho_D$ and $\rho_L$. A Spearman's correlation close to 1 indicates better performance. ``/'' indicates the baseline cannot apply to this benchmark.}
\label{tab:industrial}
\begin{tabular}{lcccccccccc}

\toprule
\multirow{2}{*}{\textbf{Model}} & \multicolumn{4}{c}{\textbf{Shape-Net Car}} & \multicolumn{4}{c}{\textbf{AirfRANS}} & \multicolumn{2}{c}{\textbf{AirCraft}} \\
\cmidrule(lr){2-5} \cmidrule(lr){6-9} \cmidrule(lr){10-11}
 & Volume $\downarrow$ & Surf $\downarrow$ & $C_D$ $\downarrow$ & $\rho_D$ $\uparrow$ & Volume $\downarrow$ & Surf $\downarrow$ & $C_L$ $\downarrow$ & $\rho_L$ $\uparrow$ & $C_L$ $\downarrow$ & Surf $\downarrow$ \\
\midrule
Simple MLP & 0.0512 & 0.1304 & 0.0307 & 0.9496 & 0.0081 & 0.0200 & 0.2108 & 0.9932 & / & / \\
GraphSAGE~\citeyearpar{sagehamilton2017inductive} & 0.0461 & 0.1050 & 0.0270 & 0.9695 & 0.0087 & 0.0184 & 0.1476 & 0.9964 & 0.040 & 0.109 \\
PointNet~\citeyearpar{qi2017pointnet} & 0.0494 & 0.1104 & 0.0298 & 0.9583 & 0.0253 & 0.0996 & 0.1973 & 0.9919 & 0.095 & 0.169 \\
Graph U-Net~\citeyearpar{graphunet2019graph} & 0.0471 & 0.1102 & 0.0226 & 0.9725 & 0.0076 & 0.0144 & 0.1677 & 0.9949 & 0.063 & 0.161 \\
MeshGraphNet~\citeyearpar{MeshGraphDBLP:conf/iclr/PfaffFSB21} & 0.0354 & 0.0781 & 0.0168 & 0.9840 & 0.0214 & 0.0387 & 0.2252 & 0.9945 & 0.038 & 0.113 \\
GNO~\citeyearpar{GNOli2020neural} & 0.0383 & 0.0815 & 0.0172 & 0.9834 & 0.0269 & 0.0405 & 0.2016 & 0.9938 & 0.031 & 0.129 \\
Galerkin~\citeyearpar{galerkinDBLP:conf/nips/Cao21} & 0.0339 & 0.0878 & 0.0179 & 0.9764 & 0.0074 & 0.0159 & 0.2336 & 0.9951 & 0.069 & 0.118 \\
Geo-FNO~\citeyearpar{geofnoli2023fourier} & 0.1670 & 0.2378 & 0.0664 & 0.8280 & 0.0361 & 0.0301 & 0.6161 & 0.9257 & 0.243 & 0.395 \\
GNOT~\citeyearpar{GNOTDBLP:conf/icml/HaoWSYDLCSZ23} & 0.0329 & 0.0798 & 0.0178 & 0.9833 & 0.0049 & 0.0152 & 0.1992 & 0.9942 & 0.033 & 0.093 \\
GINO~\citeyearpar{GINODBLP:conf/nips/LiKCLKONS0AA23} & 0.0386 & 0.0810 & 0.0184 & 0.9826 & 0.0297 & 0.0482 & 0.1821 & 0.9958 & 0.047 & 0.133 \\
3D-GeoCA~\citeyearpar{3dgeoDBLP:conf/ijcai/DengLXHM24} & 0.0319 & 0.0779 & 0.0159 & 0.9842 & / & / & / & / & 0.022 & 0.097 \\
Transolver (++)~\citeyearpar{transolverDBLP:conf/icml/WuLW0L24,trans++DBLP:conf/icml/LuoWZXD0L25} & \underline{0.0231} & 0.0787 & \underline{0.0113} & \underline{0.9901} & \underline{0.0037} & 0.0142 & 0.1030 & 0.9978 & \underline{0.023} & \underline{0.065} \\
SAOT~\citeyearpar{SAOTzhou2025dual} & 0.0223 & \underline{0.0778} & 0.0122 & 0.9884 & 0.0038 & \underline{0.0128} & \underline{0.0872} & \underline{0.9983} & 0.043 & 0.096 \\
\midrule
\cellcolor{gray!20}\textbf{PGOT (Ours)} & \cellcolor{gray!20}\textbf{0.0210} & \cellcolor{gray!20}\textbf{0.0766} & \cellcolor{gray!20}\textbf{0.0101} & \cellcolor{gray!20}\textbf{0.9926} & \cellcolor{gray!20}\textbf{0.0025} & \cellcolor{gray!20}\textbf{0.0024} & \cellcolor{gray!20}\textbf{0.0766} & \cellcolor{gray!20}\textbf{0.9990} & \cellcolor{gray!20}\textbf{0.015} & \cellcolor{gray!20}\textbf{0.044} \\
\midrule
\textbf{Improvement} & 9.1\% & 1.5\% & 10.6\% & / & 32.4\% & 81.3\% & 12.2\% & / & 34.8\% & 32.3\% \\
\bottomrule
\end{tabular}
\end{table*}
\paragraph{Standard Benchmarks.} Table~\ref{tab:standard} presents the performance comparison on four standard benchmarks. PGOT consistently achieves state-of-the-art results across all datasets, with relative error reductions ranging from 7.7\% to 12.7\% compared to the second-best methods. On Plasticity and Pipe, PGOT outperforms Transolver (++) by 7.7\% and 9.3\%, respectively, demonstrating the effectiveness of explicit geometric injection in capturing temporal dynamics and curved boundary features. For Airfoil, where shock waves introduce sharp discontinuities, PGOT achieves 9.8\% improvement over SAOT, validating the benefit of TaylorDecomp-FFN in adaptively routing computations to non-linear paths for discontinuity regions. This is qualitatively supported by Figure~\ref{fig:vis_standard}, where PGOT produces substantially cleaner predictions near shock waves compared to Transolver (++) and SAOT. The most significant improvement (12.7\%) is observed on Elasticity, a point cloud benchmark where geometric information is particularly sparse. As visualized in Figure~\ref{fig:vis_standard} (right), baselines exhibit scattered high-error points, whereas PGOT maintains consistently low errors, highlighting the importance of multi-scale geometric encoding in preserving structural details.

\begin{figure*}[htbp]
    \centering
    \includegraphics[width=1\textwidth]{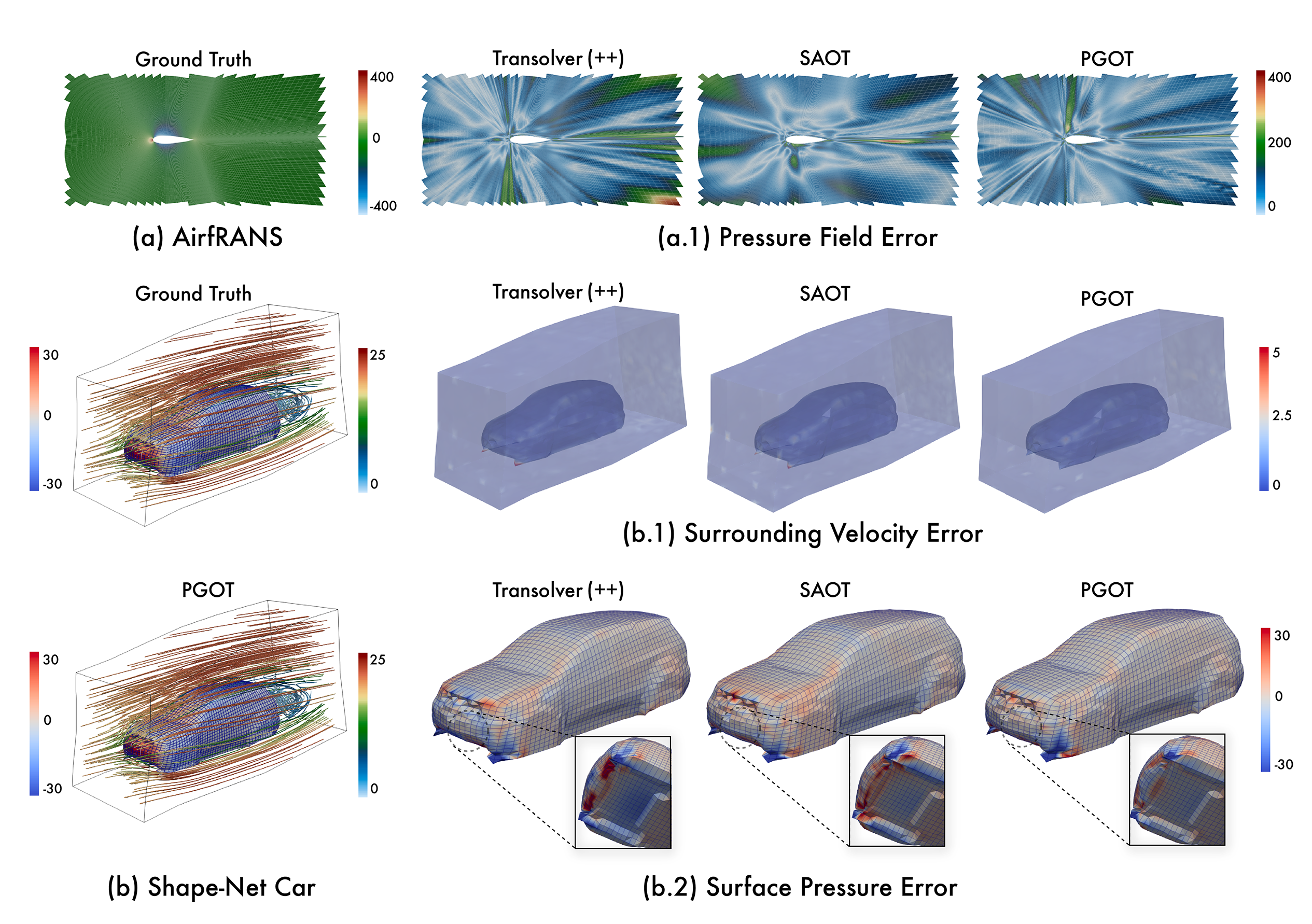}
    \caption{Visualization on industrial benchmarks. (a) AirfRANS: ground truth pressure field and prediction errors. (b) Shape-Net Car: ground truth streamlines and PGOT prediction, along with surrounding velocity and surface pressure errors.}
    \label{fig:vis_industrial}
\end{figure*}

\begin{figure}[htbp]
    \centering
    \includegraphics[width=0.4\textwidth]{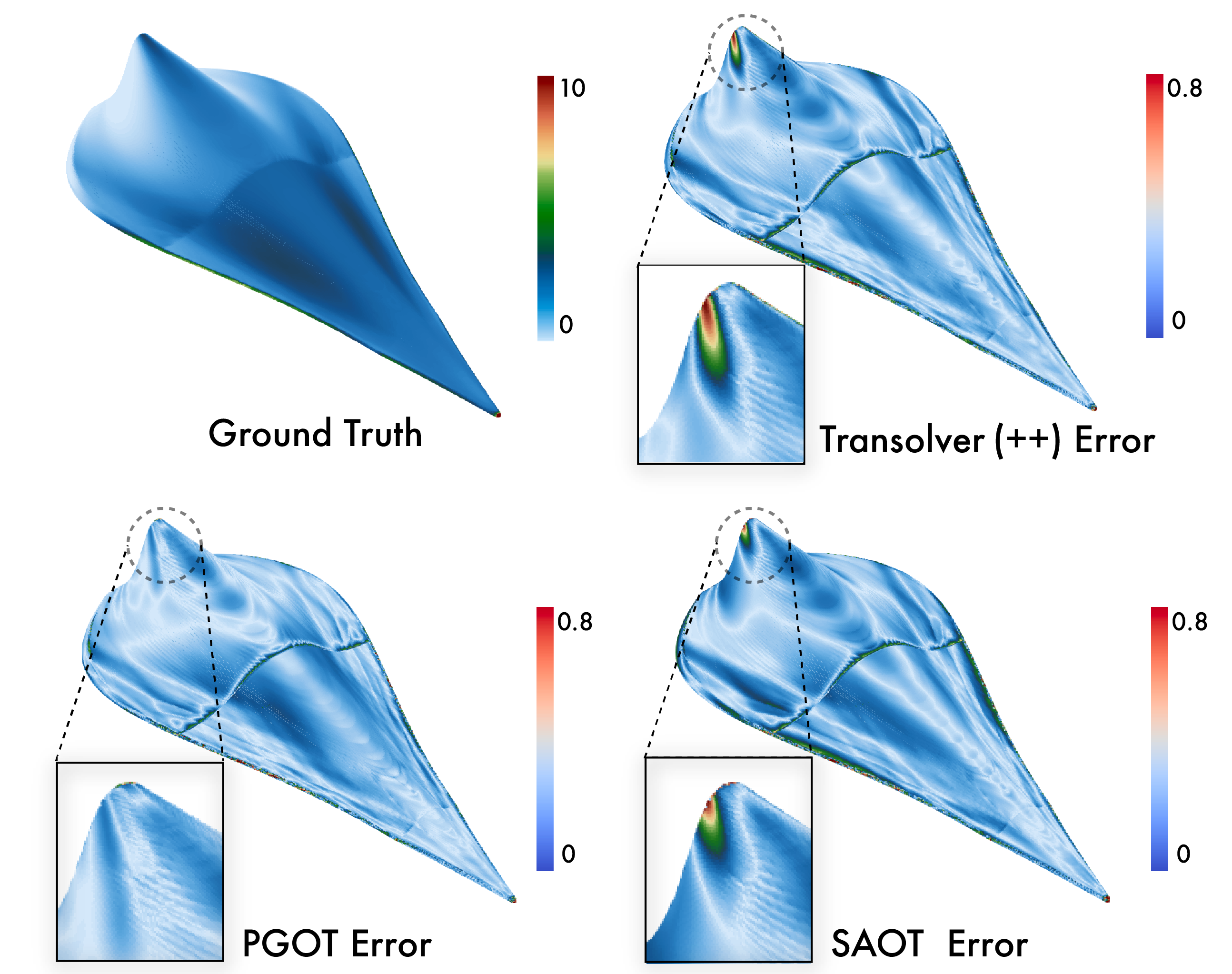}
    \caption{Visualization on the AirCraft benchmark. Ground truth surface 
    pressure field (top-left) and prediction errors for Transolver (++) 
    (top-right), PGOT (bottom-left), and SAOT (bottom-right) on a 3D wing 
    geometry.}
    \label{fig:vis_aircraft}
\end{figure}

\paragraph{Industrial Simulations.} Table~\ref{tab:industrial} summarizes 
results on large-scale industrial benchmarks. On Shape-Net Car, PGOT achieves 
the best performance across all metrics, with 9.1\% reduction in volumetric 
field error and 10.6\% improvement in drag coefficient prediction compared to 
Transolver (++). The Spearman correlation $\rho_D = 0.9926$ indicates excellent 
ranking capability for design optimization. Figure~\ref{fig:vis_industrial}(b) 
visualizes these results, showing excellent agreement between PGOT's predicted 
streamlines and ground truth, with superior accuracy in geometrically complex 
regions like wheel wells. On AirfRANS, PGOT demonstrates substantial 
improvements: 32.4\% reduction in volumetric field error, 81.3\% reduction in 
surface field error, and 12.2\% improvement in lift coefficient prediction. As 
shown in Figure~\ref{fig:vis_industrial}(a), while Transolver (++) and SAOT produce 
radial error patterns, PGOT achieves significantly cleaner predictions with 
errors primarily concentrated in the far-field region. On the AirCraft 
benchmark, PGOT achieves 34.8\% and 32.3\% reductions in $C_L$ and surface 
field errors compared to Transolver (++). As shown in 
Figure~\ref{fig:vis_aircraft}, the leading edge region---where surface 
curvature is highest and geometric aliasing is most severe---reveals a 
pronounced high-error hotspot in Transolver (++), 
which PGOT suppresses to near-zero, directly confirming the effectiveness of 
SpecGeo-Attention in high-curvature 3D geometries. These results collectively 
underscore PGOT's capability in handling complex unstructured meshes while 
maintaining high prediction fidelity. Efficiency analysis is detailed in 
Appendix~C.2.

\subsection{Ablation Studies}

We conduct systematic ablation studies to validate the contribution of each proposed component. We report results on four standard benchmarks (Plasticity, Pipe, Airfoil, Elasticity) in the main text to cover diverse geometric types and physical regimes; full results covering all datasets are provided in Appendix~C.1.

\begin{table}[htbp]
\centering
\caption{Ablation on the number of latent slices $M$. We observe that $M=64$ offers the best trade-off for standard benchmarks.}
\label{tab:ablation_slices}
    \vspace{-0.3cm}
\begin{tabular}{ccccc}
\toprule
Slices $M$ & Plasticity & Pipe & Airfoil & Elasticity \\
\midrule
8   & 0.0104 & 0.0079 & 0.0194 & 0.0239 \\
16  & 0.0047 & 0.0071 & 0.0101 & 0.0074 \\
32  & 0.0025 & 0.0040 & 0.0049 & 0.0081 \\
\cellcolor{gray!20}64  & \cellcolor{gray!20}\textbf{0.0012} & \cellcolor{gray!20}\textbf{0.0039} & \cellcolor{gray!20}\textbf{0.0046} & \cellcolor{gray!20}\underline{0.0069} \\
128 & 0.0021 & 0.0051 & 0.0056 & \textbf{0.0061} \\
256 & 0.0142 & 0.0044 & 0.0048 & 0.0079 \\
\bottomrule
\end{tabular}

\end{table}

\paragraph{Number of Latent Slices $M$.} Table~\ref{tab:ablation_slices} examines the impact of varying $M$. Performance generally improves as $M$ increases, with Plasticity, Pipe, and Airfoil peaking at $M=64$. Interestingly, Elasticity benefits from even finer granularity, reaching its best performance at $M=128$ (0.0061). This is likely because point cloud data lacks explicit connectivity, requiring more slices to capture local geometric nuances. However, further increasing $M$ to 256 leads to degradation across all tasks due to the fragmentation of coherent physical regions. Considering the trade-off between accuracy and computational cost, we adopt $M=64$ as the unified default.

\begin{table}[htbp]
\centering
\caption{Ablation on the number of geometric scales $S$. Multi-scale encoding outperforms single-scale baselines.}
    \vspace{-0.3cm}
\label{tab:ablation_scales}
\begin{tabular}{ccccc}
\toprule
Scales $S$ & Plasticity & Pipe & Airfoil & Elasticity \\
\midrule
1 & 0.0078 & 0.0068 & 0.0118 & 0.0145 \\
\cellcolor{gray!20}2 & \cellcolor{gray!20}\textbf{0.0012} & \cellcolor{gray!20}\textbf{0.0039} & \cellcolor{gray!20}\textbf{0.0046} & \cellcolor{gray!20}\textbf{0.0069} \\
3 & 0.0028 & 0.0051 & 0.0065 & 0.0083 \\
4 & 0.0042 & 0.0062 & 0.0071 & 0.0103 \\
\bottomrule
\end{tabular}
\end{table}

\paragraph{Number of Geometric Scales $S$.} Table~\ref{tab:ablation_scales} investigates the effect of multi-scale geometric encoding. Single-scale encoding ($S=1$) performs poorly across all benchmarks, confirming that aggregation-induced geometric aliasing significantly degrades prediction quality. Performance improves substantially and reaches optimality at $S=2$, demonstrating that a dual-scale representation is sufficient to capture both global topology and local boundary details. Interestingly, further increasing the scales to $S=3$ or $S=4$ leads to performance degradation. This suggests that excessive high-frequency geometric priors may introduce spectral noise or redundancy, which interferes with the model's ability to learn robust physical dynamics. See Appendix~D for additional visualizations.
\begin{table}[htbp]
\centering
\caption{Both SpecGeo-Attention (SGA) and TaylorDecomp-FFN (TDF) are critical for optimal performance.}
\label{tab:ablation_components}
    \vspace{-0.3cm}
\resizebox{\columnwidth}{!}{
\begin{tabular}{ccccc}
\toprule
Variant & Plasticity & Pipe & Airfoil & Elasticity \\
\midrule
w/o SGA & 0.0093 & 0.0045 & 0.0104 & 0.0074 \\
w/o TDF  & 0.0134 & 0.0089 & 0.0094 & 0.0231 \\
\cellcolor{gray!20}Full PGOT & \cellcolor{gray!20}\textbf{0.0012} & \cellcolor{gray!20}\textbf{0.0039} & \cellcolor{gray!20}\textbf{0.0046} & \cellcolor{gray!20}\textbf{0.0069} \\
\bottomrule
\end{tabular}
}
\end{table}

\paragraph{Component Analysis.} Table~\ref{tab:ablation_components} validates the necessity of both proposed modules. Removing SpecGeo-Attention (w/o SGA) leads to performance drops across all tasks, particularly in Elasticity ($0.0069 \to 0.0074$) and Airfoil ($0.0046 \to 0.0104$), confirming that explicit geometric injection is crucial for handling irregular point clouds and shock boundaries. Removing TaylorDecomp-FFN (w/o TDF) causes severe degradation in Plasticity ($0.0012 \to 0.0134$), validating the importance of adaptive routing for time-dependent heterogeneous fields.
\begin{table}[htbp]
\centering
\small
\caption{Analysis of FFN design choices. Standard FFN uses a single non-linear path. Standard MoE employs two identical non-linear experts with feature-based top-1 gating.}
\label{tab:ablation_ffn}
    \vspace{-0.3cm}
\begin{tabular}{ccccc}
\toprule
Variant & Plasticity & Pipe & Airfoil & Elasticity \\
\midrule
Standard FFN         & 0.0134 & 0.0089 & 0.0094 & 0.0231 \\
Standard MoE         & 0.0101 & 0.0095 & 0.0088 & 0.0215 \\
Fixed $\alpha{=}0.5$ & 0.0942 & 0.0078 & 0.0067 & 0.0103 \\
\cellcolor{gray!20}TaylorDecomp-FFN & \cellcolor{gray!20}\textbf{0.0012} & \cellcolor{gray!20}\textbf{0.0039} & \cellcolor{gray!20}\textbf{0.0046} & \cellcolor{gray!20}\textbf{0.0069} \\
\bottomrule
\end{tabular}
\end{table}
\paragraph{FFN Design Choices.} Table~\ref{tab:ablation_ffn} compares different FFN architectures with matched parameter counts. Replacing TaylorDecomp-FFN with a standard single-path FFN (Standard FFN) leads to severe degradation, confirming the benefit of dual-path computation. Standard MoE, which uses two identical non-linear experts with feature-based gating, improves over Standard FFN but remains inferior to TaylorDecomp-FFN across all benchmarks. This demonstrates that the explicit decomposition into linear and non-linear expert paths---aligned with the smooth and discontinuous components of physical fields---is more effective than homogeneous expert routing. Furthermore, fixing $\alpha{=}0.5$ uniformly across the domain degrades performance compared to the full spatially adaptive gate, confirming that geometry-aware routing is essential for allocating computational capacity to regions of varying physical complexity.
\subsection{Out-of-Distribution (OOD) Generalization}
\label{sec:ood}

We evaluate PGOT on AirfRANS under two extrapolation settings: (1) Unseen Reynolds and (2) Unseen Angles. As shown in Table~\ref{tab:ood_main}, PGOT demonstrates superior robustness against 11 baselines. In Reynolds extrapolation, PGOT achieves the lowest lift coefficient error ($C_L=0.2942$) and the highest ranking correlation ($\rho_L=0.9923$). In Angle extrapolation, PGOT maintains the highest ranking consistency ($\rho_L=0.9952$), significantly outperforming graph-based methods and surpassing the SOTA Transolver (++). This confirms that explicitly modeling geometry-physics interactions enables robust reasoning in unseen physical regimes. Detailed analysis are provided in Appendix~C.3.

\begin{table}[htbp]
\centering
\caption{OOD generalization on AirfRANS.}
\label{tab:ood_main}
    \vspace{-0.3cm}
\resizebox{\columnwidth}{!}{
\begin{tabular}{lcccc}
\toprule
\multirow{2}{*}{\textbf{Model}} & \multicolumn{2}{c}{\textbf{Unseen Reynolds}} & \multicolumn{2}{c}{\textbf{Unseen Angles}} \\
\cmidrule(lr){2-3} \cmidrule(lr){4-5}
 & $C_L$ $\downarrow$ & $\rho_L$ $\uparrow$ & $C_L$ $\downarrow$ & $\rho_L $$\uparrow$ \\
\midrule
Simple MLP & 0.6205 & 0.9578 & 0.4128 & 0.9572 \\
GraphSAGE~\citeyearpar{sagehamilton2017inductive} & 0.4333 & 0.9707 & 0.2538 & 0.9894 \\
PointNet~\citeyearpar{qi2017pointnet} & 0.3836 & 0.9806 & 0.4425 & 0.9784 \\
Graph U-Net~\citeyearpar{graphunet2019graph} & 0.4664 & 0.9645 & 0.3756 & 0.9816 \\
MeshGraphNet~\citeyearpar{MeshGraphDBLP:conf/iclr/PfaffFSB21} & 1.7718 & 0.7631 & 0.6525 & 0.8927 \\
GNO~\citeyearpar{GNOli2020neural} & 0.4408 & 0.9878 & 0.3038 & 0.9884 \\
Galerkin~\citeyearpar{galerkinDBLP:conf/nips/Cao21} & 0.4615 & 0.9826 & 0.3814 & 0.9821 \\
GNOT~\citeyearpar{GNOTDBLP:conf/icml/HaoWSYDLCSZ23} & 0.3268 & 0.9865 & 0.3497 & 0.9868 \\
GINO~\citeyearpar{GINODBLP:conf/nips/LiKCLKONS0AA23} & 0.4180 & 0.9645 & 0.2583 & 0.9923 \\
Transolver (++)~\citeyearpar{transolverDBLP:conf/icml/WuLW0L24,trans++DBLP:conf/icml/LuoWZXD0L25} & \underline{0.3042} & \underline{0.9899} & \textbf{0.1503} & \underline{0.9948} \\
SAOT~\citeyearpar{SAOTzhou2025dual} & 0.4242 & 0.9764 & 0.2925 & 0.9888 \\
\midrule
\cellcolor{gray!20}\textbf{PGOT (Ours)} & \cellcolor{gray!20}\textbf{0.2942} & \cellcolor{gray!20}\textbf{0.9923} & \cellcolor{gray!20}\underline{0.1742} & \cellcolor{gray!20}\textbf{0.9952} \\
\bottomrule
\end{tabular}
}
    \vspace{-0.3cm}
\end{table}

\section{Limitations}
While PGOT achieves strong performance across diverse benchmarks,
the multi-scale geometric encoding introduces additional
computational overhead that may become non-trivial for meshes
with extremely irregular point distributions.
Additionally, the exponential scaling strategy relies on
manually chosen scale factors, and exploring adaptive scale
learning to reduce such tuning remains an open direction.
Future work may also extend the geometric-physical decoupling
paradigm to multi-physics foundation models.

\section{Conclusion}
\label{sec:conclusion}

In this work, we proposed PGOT to address the geometric aliasing bottleneck in efficient neural operators. By synergizing Spectrum-Preserving Geometric Attention with a regime-adaptive Taylor-Decomposed FFN, our framework reconciles linear complexity with high-fidelity boundary reconstruction. PGOT achieves state-of-the-art results on industrial benchmarks and demonstrates superior robustness in extrapolation tasks. We hope this work inspires future research on geometry-aware operator learning as a principled foundation for high-fidelity scientific computing.

\bibliography{example_paper}
\bibliographystyle{icml2026}

\newpage
\appendix
\onecolumn
\renewcommand{\thesection}{Appendix \Alph{section}}

\section{Implementation Details}
\label{app:implementation}
\begin{figure*}[h]
    \centering
    \includegraphics[width=1\textwidth]{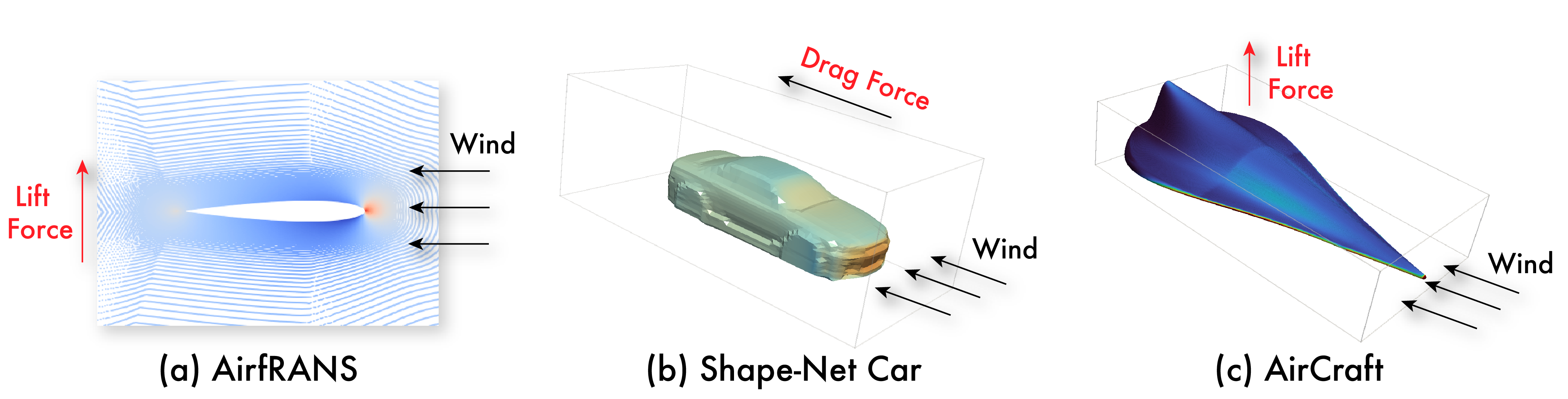}
    \caption{Industrial design tasks for aerodynamic simulation, 
    covering airfoil, car, and aircraft geometries.}
    \label{fig:design_tasks}
\end{figure*}
\subsection{Dataset Descriptions}
\label{app:datasets}

\begin{figure*}[htbp]
    \centering
    \includegraphics[width=0.9\textwidth]{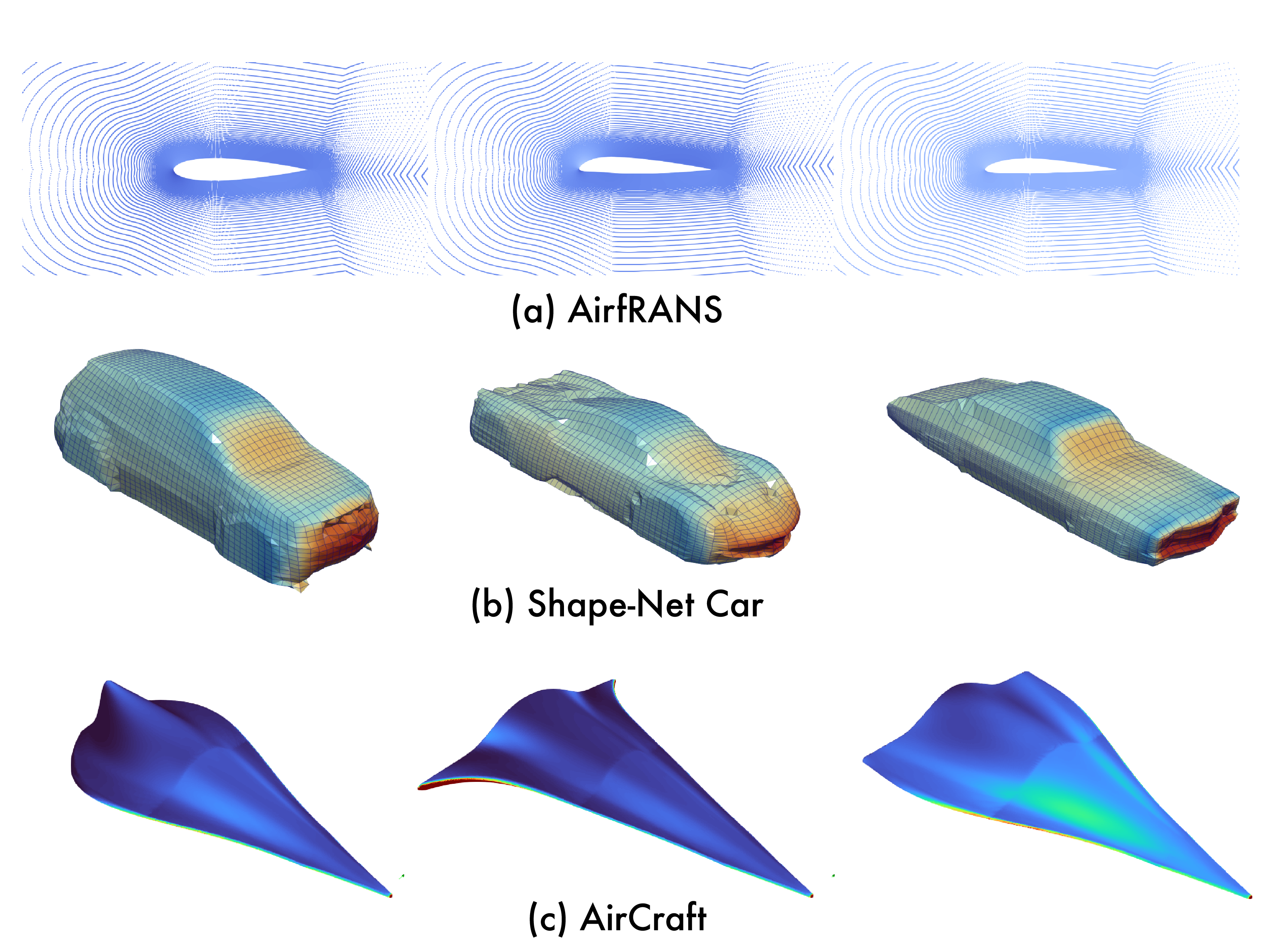}
    \caption{Visualization of representative samples from the 
    three industrial benchmarks.}
    \label{fig:dataset_examples}
\end{figure*}

We provide detailed descriptions of all seven benchmarks used in our evaluation. Table~\ref{tab:dataset_full} summarizes the key statistics. The three industrial benchmarks and their representative 
samples are illustrated in Figures~\ref{fig:design_tasks} 
and~\ref{fig:dataset_examples}.

\begin{table*}[htbp]
\centering
\caption{Summary of benchmark datasets. \#Mesh denotes the number of discretized mesh points per sample.}
\label{tab:dataset_full}
\begin{tabular}{llccccc}
\toprule
\textbf{Geometry} & \textbf{Benchmark} & \textbf{Dim} & \textbf{\#Mesh} & \textbf{Input} & \textbf{Output} & \textbf{Train / Test} \\
\midrule
Point Cloud & Elasticity & 2D & 972 & Structure & Inner Stress & 1000 / 200 \\
\multirow{3}{*}{Structured Mesh} & Plasticity & 2D+T & 3,131 & External Force & Displacement & 900 / 80 \\
 & Airfoil & 2D & 11,271 & Structure & Mach Number & 1000 / 200 \\
 & Pipe & 2D & 16,641 & Structure & Velocity & 1000 / 200 \\
\midrule
\multirow{3}{*}{Unstructured Mesh} & Shape-Net Car & 3D & 32,186 & Structure & Velocity, Pressure & 789 / 100 \\
 & AirfRANS & 2D & 32,000 & Structure & Velocity, Pressure & 800 / 200 \\
 & AirCraft & 3D & $\sim$300K & Structure & 6 Quantities & 140 / 10 \\
\bottomrule
\end{tabular}
\end{table*}

\begin{table*}[htbp]
\centering
\caption{Training configurations for each benchmark. All methods are trained under identical settings.}
\label{tab:training_config}
\begin{tabular}{lccccc}
\toprule
\textbf{Benchmark} & \textbf{Loss} & \textbf{Epochs} & \textbf{Learning Rate} & \textbf{Optimizer} & \textbf{Batch Size} \\
\midrule
Elasticity & Relative $L^2$ & 500 & $10^{-3}$ & AdamW & 1 \\
Plasticity & Relative $L^2$ & 500 & $10^{-3}$ & AdamW & 8 \\
Airfoil & Relative $L^2$ & 500 & $10^{-3}$ & AdamW & 4 \\
Pipe & Relative $L^2$ & 500 & $10^{-3}$ & AdamW & 4 \\
Shape-Net Car & $\mathcal{L}_v + 0.5\mathcal{L}_s$ & 200 & $10^{-3}$ & Adam & 1 \\
AirfRANS & $\mathcal{L}_v + \mathcal{L}_s$ & 400 & $10^{-3}$ & Adam & 1 \\
AirCraft & $\mathcal{L}_v + \mathcal{L}_s$ & 400 & $10^{-3}$ & Adam & 1 \\
\bottomrule
\end{tabular}
\end{table*}

\begin{table*}[htbp]
\centering
\caption{PGOT model configurations for each benchmark.}
\label{tab:model_config}
\begin{tabular}{lccccc}
\toprule
\textbf{Benchmark} & \textbf{Layers $L$} & \textbf{Channels $C$} & \textbf{Slices $M$} & \textbf{Scales $S$} & \textbf{Heads} \\
\midrule
Elasticity & 8  & 128 & 64 & 2 & 8 \\
Plasticity & 8 & 128 & 64 & 2 & 8 \\
Airfoil & 8 & 128 & 64 & 2 & 8 \\
Pipe & 8 & 128 & 64 & 2 & 8 \\
Shape-Net Car & 8 & 256 & 32 & 3 & 8 \\
AirfRANS & 8 & 256 & 32 & 3 & 8 \\
AirCraft & 8 & 256 & 32 & 3 & 8 \\
\bottomrule
\end{tabular}
\end{table*}

\begin{table*}[htbp]
\centering
\caption{Full ablation on the number of latent slices $M$. For standard benchmarks (Plasticity, Pipe, Airfoil, Elasticity), we fix $S=2$; for industrial benchmarks (Car, AirfRANS, AirCraft), we fix $S=3$, matching the deployed configuration of PGOT on each task type.}
\label{tab:app_ablation_slices}
\begin{tabular}{cccccccc}
\toprule
$M$ & Plasticity & Pipe & Airfoil & Elasticity & Car (Vol.) & AirfRANS (Vol.)&AirCraft (Surf.) \\
\midrule
8   & 0.0104 & 0.0079 & 0.0194 & 0.0239 & 0.0421 & 0.0077& 0.091\\
16  & 0.0047 & 0.0071 & 0.0101 & 0.0074 & 0.0394 & 0.0059&0.078\\
32  & 0.0025 & 0.0040 & 0.0049 & 0.0081 & \textbf{0.0210} & \textbf{0.0025} &\textbf{0.044}\\
64  & \textbf{0.0012} & \textbf{0.0039} & \textbf{0.0046} & 0.0069 & 0.0299 & 0.0055& 0.059\\
128 & 0.0021 & 0.0051 & 0.0056 & \textbf{0.0061} & 0.0322 & 0.0049& 0.066\\
256 & 0.0142 & 0.0044 & 0.0048 & 0.0079 & 0.0318 & 0.0104& 0.064\\
\bottomrule
\end{tabular}
\end{table*}

\begin{table*}[htbp]
\centering
\caption{Full ablation on the number of geometric scales $S$. For standard benchmarks, we fix $M=64$; for industrial benchmarks, we fix $M=32$, matching the deployed configuration. Single-scale encoding ($S=1$) corresponds to the absence of multi-scale geometric injection.}
\label{tab:app_ablation_scales}
\begin{tabular}{cccccccc}
\toprule
$S$ & Plasticity & Pipe & Airfoil & Elasticity & Car (Vol.) & AirfRANS (Vol.) & AirCraft (Surf.) \\
\midrule
1 & 0.0078 & 0.0068 & 0.0118 & 0.0145 & 0.0342 & 0.0058 & 0.092 \\
2 & \textbf{0.0012} & \textbf{0.0039} & \textbf{0.0046} & \textbf{0.0069} & 0.0237 & 0.0031 & 0.065 \\
3 & 0.0028 & 0.0051 & 0.0065 & 0.0083 & \textbf{0.0210} & \textbf{0.0025} & \textbf{0.044} \\
4 & 0.0042 & 0.0062 & 0.0071 & 0.0103 & 0.0276 & 0.0042 & 0.058 \\
\bottomrule
\end{tabular}
\end{table*}

\begin{table*}[htbp]
\centering
\caption{Full ablation on key components.}
\label{tab:app_ablation_components}
\begin{tabular}{lccccccc}
\toprule
Variant & Plasticity & Pipe & Airfoil & Elasticity & Car (Vol.) & AirfRANS (Vol.)&AirCraft (Surf.)  \\
\midrule
w/o SGA & 0.0093 & 0.0045 & 0.0104 & 0.0074 & 0.0482 & 0.0081&0.101 \\
w/o TDF  & 0.0134 & 0.0089 & 0.0094 & 0.0231 & 0.0348 & 0.0076 &0.052\\
\cellcolor{gray!20}Full PGOT & \cellcolor{gray!20}\textbf{0.0012} & \cellcolor{gray!20}\textbf{0.0039} & \cellcolor{gray!20}\textbf{0.0046} & \cellcolor{gray!20}\textbf{0.0069} & \cellcolor{gray!20}\textbf{0.0210} & \cellcolor{gray!20}\textbf{0.0025}&
\cellcolor{gray!20}\textbf{0.044}\\
\bottomrule
\end{tabular}
\end{table*}

\paragraph{Elasticity.} This benchmark involves predicting the inner stress distribution of elastic materials based on geometric structures, discretized as 972 scattered points in 2D space. Each sample represents a different material structure, and the model must estimate the stress tensor at each point. The training set contains 1000 samples, and the test set contains 200 samples~\cite{fnoDBLP:conf/iclr/LiKALBSA21, transolverDBLP:conf/icml/WuLW0L24,trans++DBLP:conf/icml/LuoWZXD0L25}.

\paragraph{Plasticity.} The task is to predict future deformations of plastic materials under impact from arbitrarily-shaped dies. The input consists of structured meshes with $101 \times 31$ points representing the die geometry, and the output is the displacement field over 20 future time steps containing four directional deformation components. The training set contains 900 samples, and the test set contains 80 samples~\cite{fnoDBLP:conf/iclr/LiKALBSA21, transolverDBLP:conf/icml/WuLW0L24,trans++DBLP:conf/icml/LuoWZXD0L25}.

\paragraph{Airfoil.} This benchmark requires estimating Mach number distributions around airfoil geometries on structured meshes with $221 \times 51$ points. All shapes are deformations of the NACA-0012 baseline airfoil. The task involves transonic flow regimes where shock waves introduce discontinuities in the solution field. The training set contains 1000 samples, and the test set contains 200 samples~\cite{fnoDBLP:conf/iclr/LiKALBSA21, transolverDBLP:conf/icml/WuLW0L24,trans++DBLP:conf/icml/LuoWZXD0L25}.

\paragraph{Pipe.} The objective is to predict horizontal fluid velocity fields based on curved pipe structures. Each sample is discretized into a $129 \times 129$ structured mesh, with different samples generated by varying the pipe centerline geometry. The training set contains 1000 samples, and the test set contains 200 samples~\cite{fnoDBLP:conf/iclr/LiKALBSA21, transolverDBLP:conf/icml/WuLW0L24,trans++DBLP:conf/icml/LuoWZXD0L25}.

\paragraph{Shape-Net Car.} This industrial benchmark focuses on automotive aerodynamic simulation at 72 km/h driving conditions. The task requires simultaneous prediction of surface pressure and surrounding air velocity. Vehicle geometries are sourced from the ``car" category of the ShapeNet repository and discretized into unstructured meshes with approximately 32,000 points combining surface and volumetric elements. The input is preprocessed to include mesh point positions, signed distance functions, and normal vectors. The predicted fields can be used to compute drag coefficients for design optimization. The training set contains 789 samples, and the test set contains 100 samples~\cite{CARumetani2018learning}.

\paragraph{AirfRANS.} This dataset contains high-fidelity Reynolds-Averaged Navier-Stokes simulation data for airfoils from the NACA 4- and 5-digit series. Unlike the Airfoil benchmark, AirfRANS features diverse airfoil shapes under varying Reynolds numbers ($3 \times 10^6$ to $6 \times 10^6$) and angles of attack ($-5°$ to $15°$), with finer unstructured meshes of 32,000 points per sample. Air velocity, pressure, and viscosity are recorded for surrounding regions, along with surface pressure. Both lift and drag coefficients can be computed from the predicted fields. However, as noted in the original paper, air velocity estimation remains challenging for aircraft, causing all deep models to perform poorly on drag coefficient prediction. Therefore, we focus primarily on lift coefficient and volume/surface physical quantity prediction accuracy. The training set contains 800 samples, and the test set contains 200 samples~\cite{AIRANSbonnet2022airfrans}.
\paragraph{AirCraft.} This benchmark comprises high-fidelity CFD simulations 
of over 30 aircraft wing designs under five different inflow conditions, 
varying in Mach number, angle of attack, and sideslip angle. Each sample 
is discretized into approximately 300,000 unstructured 3D mesh points, 
with inputs consisting of spatial coordinates $(x, y, z)$ and surface normal 
vectors. The model is required to predict six physical quantities: pressure 
coefficient $C_p$, fluid density $\rho$, velocity components $(u, v, w)$, 
and pressure $p$. These simulations are generated by aerodynamicists using 
industrial CFD solvers, ensuring high physical fidelity. The dataset contains 
140 training samples and 10 test samples, making it a challenging benchmark 
for large-scale 3D aerodynamic modeling~\cite{trans++DBLP:conf/icml/LuoWZXD0L25}.
\subsection{Evaluation Metrics}
\label{app:metrics}

\paragraph{Relative $L^2$ Error.} Given the ground-truth physical field $\mathbf{u}$ and the model prediction $\hat{\mathbf{u}}$, the relative $L^2$ error is defined as:
\begin{equation}
    \text{Relative } L^2 = \frac{\|\mathbf{u} - \hat{\mathbf{u}}\|_2}{\|\mathbf{u}\|_2}.
\end{equation}
This metric is used across all benchmarks to measure the accuracy of predicted physical fields.

\paragraph{Aerodynamic Coefficient Error.} For Shape-Net Car, AirfRANS, and AirCraft, we compute aerodynamic coefficients from the predicted fields and report their relative errors. For unit density fluid, the coefficient (drag or lift) is defined as:
\begin{equation}
    C = \frac{2}{v^2 A} \left[ \int_{\partial\Omega} p(\xi) \left( \hat{\mathbf{n}}(\xi) \cdot \hat{\mathbf{i}}(\xi) \right) d\xi + \int_{\partial\Omega} \tau(\xi) \cdot \hat{\mathbf{i}}(\xi) d\xi \right],
\end{equation}
where $v$ is the inlet flow speed, $A$ is the reference area, $\partial\Omega$ denotes the object surface, $p$ is the pressure function, $\hat{\mathbf{n}}$ is the outward unit normal vector of the surface, $\hat{\mathbf{i}}$ is the inlet flow direction, and $\tau$ represents the wall shear stress on the surface. The wall shear stress $\tau$ can be calculated from the air velocity near the surface, which is typically much smaller than the pressure term. For the drag coefficient of Shape-Net Car, $\hat{\mathbf{i}} = (-1, 0, 0)$ and $A$ is the area of the smallest rectangle enclosing the front of the car. For the lift coefficient of AirfRANS, $\hat{\mathbf{i}} = (0, 0, -1)$. The relative $L^2$ error is computed between the ground-truth coefficient and the coefficient calculated from the predicted fields.

\paragraph{Spearman's Rank Correlation.} Given $K$ test samples with ground-truth coefficients $C = \{C^1, \cdots, C^K\}$ (drag or lift) and model-predicted coefficients $\hat{C} = \{\hat{C}^1, \cdots, \hat{C}^K\}$, Spearman's correlation coefficient is defined as the Pearson correlation coefficient between the rank variables:
\begin{equation}
    \rho = \frac{\text{cov}(R(C), R(\hat{C}))}{\sigma_{R(C)} \sigma_{R(\hat{C})}},
\end{equation}
where $R$ is the ranking function, $\text{cov}$ denotes the covariance, and $\sigma$ represents the standard deviation of the rank variables. This metric measures the correlation between the ranking distribution of ground-truth coefficients and model-predicted values across all test samples, quantifying the model's ability to correctly rank different designs. This property is particularly important for shape optimization applications. A Spearman's correlation close to 1 indicates better performance.

\subsection{Training and Model Configurations}
\label{app:training}

Following the conventions established in prior work~\citep{fnoDBLP:conf/iclr/LiKALBSA21, geofnoli2023fourier, transolverDBLP:conf/icml/WuLW0L24,SAOTzhou2025dual,trans++DBLP:conf/icml/LuoWZXD0L25}, our model is trained using the relative $L^2$ loss. Table~\ref{tab:training_config} summarizes the training configurations shared across all methods for fair comparison, and Table~\ref{tab:model_config} presents PGOT-specific hyperparameters.

For Shape-Net Car~\cite{CARumetani2018learning}, AirfRANS~\cite{AIRANSbonnet2022airfrans}, and AirCraft~\cite{trans++DBLP:conf/icml/LuoWZXD0L25}, 
$\mathcal{L}_v$ and $\mathcal{L}_s$ denote the relative $L^2$ losses on volumetric and surface fields.

The hidden dimension $C$ is set to 256 for high-dimensional output tasks (Shape-Net Car, AirfRANS, and AirCraft) and 128 for others. Accordingly, the number of latent slices $M$ is adjusted to 32 for $C=256$ and 64 for $C=128$ to balance model capacity and computational efficiency. These settings are consistent with the configurations used in Transolver (++)~\cite{transolverDBLP:conf/icml/WuLW0L24,trans++DBLP:conf/icml/LuoWZXD0L25} and other neural operators, and their effectiveness is further validated by our ablation studies.
\subsection{Baseline Implementations}
\label{app:baselines}

\textbf{Neural Operators.} For FNO~\cite{fnoDBLP:conf/iclr/LiKALBSA21}, Geo-FNO~\cite{geofnoli2023fourier}, WMT~\cite{MWTDBLP:conf/nips/GuptaXB21}, U-FNO~\cite{unfnowen2022u}, U-NO~\cite{unoDBLP:journals/tmlr/RahmanRA23}, F-FNO~\cite{ffnoDBLP:conf/iclr/TranMXO23}, and LSM~\cite{LSMDBLP:conf/icml/WuHLWL23}, we report results from the original publications~\cite{transolverDBLP:conf/icml/WuLW0L24,SAOTzhou2025dual,fnoDBLP:conf/iclr/LiKALBSA21,trans++DBLP:conf/icml/LuoWZXD0L25}.

\noindent\textbf{Transformer-based Methods.} For GNOT~\cite{GNOTDBLP:conf/icml/HaoWSYDLCSZ23}, ONO~\cite{ONODBLP:conf/icml/XiaoHLD024}, Galerkin Transformer~\cite{galerkinDBLP:conf/nips/Cao21}, HT-Net~\cite{htnetliu2024mitigating}, FactFormer~\cite{FactFormerDBLP:conf/nips/LiSF23}, and OFormer~\cite{oformerDBLP:journals/tmlr/LiMF23}, we report results from the original publications~\cite{transolverDBLP:conf/icml/WuLW0L24,SAOTzhou2025dual,fnoDBLP:conf/iclr/LiKALBSA21}. For Transolver (++)~\cite{transolverDBLP:conf/icml/WuLW0L24,trans++DBLP:conf/icml/LuoWZXD0L25} and SAOT~\cite{SAOTzhou2025dual}, we reproduce results based on their official codebases with all hyperparameter configurations kept consistent with PGOT to ensure fair comparison.

\noindent\textbf{Geometric Deep Learning Methods.} For MLP, GraphSAGE~\cite{sagehamilton2017inductive}, PointNet~\cite{qi2017pointnet}, Graph U-Net~\cite{graphunet2019graph}, MeshGraphNet~\cite{MeshGraphDBLP:conf/iclr/PfaffFSB21}, GNO~\cite{GNOli2020neural}, GINO~\cite{GINODBLP:conf/nips/LiKCLKONS0AA23}, and 3D-GeoCA~\cite{3dgeoDBLP:conf/ijcai/DengLXHM24}, we report results from the original publications.

\section{Supplementary Results}
\begin{table*}[htbp]
\centering
\caption{Efficiency comparison of Transformer-based methods across different mesh sizes $N$. Running time is measured by the time (s) to complete one epoch containing 1000 iterations. Params denotes the number of model parameters (MB).}
\label{tab:app_efficiency}
\resizebox{\textwidth}{!}{
\begin{tabular}{lc|cccccc|cccccc}
\toprule
\multirow{2}{*}{\textbf{Model}} & \multirow{2}{*}{\textbf{Params}} & \multicolumn{6}{c|}{\textbf{GPU Memory (GB)}} & \multicolumn{6}{c}{\textbf{Time (s/epoch)}} \\
\cmidrule(lr){3-8} \cmidrule(lr){9-14}
& & 1024 & 2048 & 4096 & 8192 & 16384 & 32768 & 1024 & 2048 & 4096 & 8192 & 16384 & 32768 \\
\midrule
Galerkin~\citeyearpar{galerkinDBLP:conf/nips/Cao21} & 1.041 & 0.62 & 0.66 & 0.74 & 0.91 & 1.45 & 2.05 & 26.51 & 26.50 & 27.48 & 37.10 & 67.52 & 129.87 \\
OFormer~\citeyearpar{oformerDBLP:journals/tmlr/LiMF23} & 0.884 & 0.63 & 0.69 & 0.80 & 1.02 & 1.67 & 2.44 & 28.15 & 30.98 & 31.11 & 47.90 & 91.67 & 182.21 \\
ONO~\citeyearpar{ONODBLP:conf/icml/XiaoHLD024} & 1.109 & 1.47 & 1.75 & 2.30 & 3.47 & 5.64 & 10.09 & 69.76 & 76.25 & 100.13 & 149.59 & 255.34 & 462.46 \\
GNOT~\citeyearpar{GNOTDBLP:conf/icml/HaoWSYDLCSZ23} & 5.248 & 0.85 & 1.07 & 1.47 & 2.33 & 4.23 & 7.46 & 54.28 & 55.94 & 60.86 & 67.17 & 112.55 & 209.92 \\
Transolver (++)~\citeyearpar{transolverDBLP:conf/icml/WuLW0L24,trans++DBLP:conf/icml/LuoWZXD0L25} & 0.950 & 0.15 & 0.27 & 0.53 & 1.02 & 2.03 & 4.02 & 38.43 & 38.56 & 37.33 & 38.36 & 51.73 & 98.54 \\
SAOT~\citeyearpar{SAOTzhou2025dual} & 1.492 & 0.33 & 0.52 & 0.92 & 1.61 & 3.11 & 6.33 & 94.09 & 94.72 & 97.01 & 97.22 & 132.33 & 234.94 \\
\midrule
\cellcolor{gray!20}\textbf{PGOT (Ours)} & \cellcolor{gray!20}1.148 & \cellcolor{gray!20}0.20 & \cellcolor{gray!20}0.37 & \cellcolor{gray!20}0.71 & \cellcolor{gray!20}1.38 & \cellcolor{gray!20}2.71 & \cellcolor{gray!20}5.37 & \cellcolor{gray!20}58.04 & \cellcolor{gray!20}59.76 & \cellcolor{gray!20}59.52 & \cellcolor{gray!20}60.44 & \cellcolor{gray!20}60.17 & \cellcolor{gray!20}110.99 \\
\bottomrule
\end{tabular}
}
\end{table*}

\begin{table*}[h]
\caption{Settings of OOD generalization experiments on AirfRANS. The range of Reynolds numbers and angles of attack are listed.}
\label{tab:ood_settings}
\centering
\begin{tabular}{l|cc|cc}
\toprule
\multirow{2}{*}{Dataset} & \multicolumn{2}{c|}{Unseen Reynolds} & \multicolumn{2}{c}{Unseen Angles} \\
& Reynolds Range & Samples & Angles Range & Samples \\
\midrule
Training Set & $[3\times 10^6, 5\times 10^6]$ & 500 & $[-2.5°, 12.5°]$ & 800 \\
Test Set & $[2\times 10^6, 3\times 10^6] \cup [5\times 10^6, 6\times 10^6]$ & 500 & $[-5°, -2.5°] \cup [12.5°, 15°]$ & 200 \\
\bottomrule
\end{tabular}
\end{table*}

\begin{table*}[htbp]
\centering
\caption{Full OOD generalization results on AirfRANS. PGOT achieves the best overall performance, particularly in ranking consistency ($\rho_L$) and field reconstruction (Volume/Surf). Best results are \textbf{bolded}, second best are \underline{underlined}.}
\label{tab:app_ood_full}
\resizebox{\textwidth}{!}{
\begin{tabular}{lcccccccc}
\toprule
\multirow{2}{*}{\textbf{Model}} & \multicolumn{4}{c}{\textbf{Unseen Reynolds Numbers}} & \multicolumn{4}{c}{\textbf{Unseen Angles of Attack}} \\
\cmidrule(lr){2-5} \cmidrule(lr){6-9}
 & Volume $\downarrow$ & Surf $\downarrow$ & $C_L$ $\downarrow$ & $\rho_L$ $\uparrow$ & Volume $\downarrow$ & Surf $\downarrow$ & $C_L$ $\downarrow$ & $\rho_L$ $\uparrow$ \\
\midrule
Simple MLP & 0.0669 & 0.1153 & 0.6205 & 0.9578 & 0.1309 & 0.3311 & 0.4128 & 0.9572 \\
GraphSAGE~\citeyearpar{sagehamilton2017inductive} & 0.0798 & 0.1254 & 0.4333 & 0.9707 & 0.1192 & 0.2359 & 0.2538 & 0.9894 \\
PointNet~\citeyearpar{qi2017pointnet} & 0.0838 & 0.1403 & 0.3836 & 0.9806 & 0.2021 & 0.4649 & 0.4425 & 0.9784 \\
Graph U-Net~\citeyearpar{graphunet2019graph} & 0.0538 & 0.1168 & 0.4664 & 0.9645 & 0.0979 & 0.2391 & 0.3756 & 0.9816 \\
MeshGraphNet~\citeyearpar{MeshGraphDBLP:conf/iclr/PfaffFSB21} & 0.2789 & 0.2382 & 1.7718 & 0.7631 & 0.4902 & 1.1071 & 0.6525 & 0.8927 \\
GNO~\citeyearpar{GNOli2020neural} & 0.0833 & 0.1562 & 0.4408 & 0.9878 & 0.1626 & 0.2359 & 0.3038 & 0.9884 \\
Galerkin~\citeyearpar{galerkinDBLP:conf/nips/Cao21} & 0.0330 & 0.0972 & 0.4615 & 0.9826 & 0.0577 & 0.2773 & 0.3814 & 0.9821 \\
GNOT~\citeyearpar{GNOTDBLP:conf/icml/HaoWSYDLCSZ23} & 0.0305 & 0.0959 & 0.3268 & 0.9865 & 0.0471 & 0.3466 & 0.3497 & 0.9868 \\
GINO~\citeyearpar{GINODBLP:conf/nips/LiKCLKONS0AA23} & 0.0839 & 0.1825 & 0.4180 & 0.9645 & 0.1589 & 0.2469 & 0.2583 & 0.9923 \\
Transolver (++)~\citeyearpar{transolverDBLP:conf/icml/WuLW0L24,trans++DBLP:conf/icml/LuoWZXD0L25} & \underline{0.0234} & \underline{0.0341} & \underline{0.3042} & \underline{0.9899} & \underline{0.0454} & \underline{0.2279} & \textbf{0.1503} & \underline{0.9948} \\
SAOT~\citeyearpar{SAOTzhou2025dual} & 0.1032 & 0.1193 & 0.4242 & 0.9764 & 0.0748 & 0.2744 & 0.2925 & 0.9888 \\
\midrule
\cellcolor{gray!20}\textbf{PGOT (Ours)} & \cellcolor{gray!20}\textbf{0.0203} & \cellcolor{gray!20}\textbf{0.0293} & \cellcolor{gray!20}\textbf{0.2942} & \cellcolor{gray!20}\textbf{0.9923} & \cellcolor{gray!20}\textbf{0.0348} & \cellcolor{gray!20}\textbf{0.2203} & \cellcolor{gray!20}\underline{0.1742} & \cellcolor{gray!20}\textbf{0.9952} \\
\bottomrule
\end{tabular}
}
\end{table*}

\subsection{Ablation Results}
\label{app:ablation}
We provide the complete ablation study results on all seven benchmarks in Tables~\ref{tab:app_ablation_slices}, \ref{tab:app_ablation_scales}, and \ref{tab:app_ablation_components}.

\subsection{Efficiency Analysis}
\label{app:efficiency}

Table~\ref{tab:app_efficiency} compares computational efficiency across different mesh sizes. PGOT maintains competitive efficiency while achieving superior accuracy. Compared to Transolver (++), PGOT requires slightly more GPU memory (5.37 GB vs. 4.02 GB at $N=32768$) but achieves substantially better prediction quality. Notably, PGOT exhibits excellent scalability: the running time remains nearly constant (58-60 s/epoch) as mesh size increases from 1024 to 16384, confirming the $\mathcal{O}(N)$ linear complexity of SpecGeo-Attention. In contrast, methods like ONO~\cite{ONODBLP:conf/icml/XiaoHLD024} and OFormer~\cite{oformerDBLP:journals/tmlr/LiMF23} show significant time increases with mesh size due to their higher computational complexity. At $N=32768$, PGOT is 4.2$\times$ faster than ONO and 1.6$\times$ faster than OFormer~\cite{oformerDBLP:journals/tmlr/LiMF23} while achieving notably lower prediction errors.

\subsection{Out-of-Distribution Generalization}
\label{app:ood}

We evaluate PGOT under two extrapolation settings on AirfRANS: (1) Unseen Reynolds numbers and (2) Unseen angles of attack. The detailed experimental settings are provided in Table~\ref{tab:ood_settings}, where the training and test sets contain completely different Reynolds numbers or angles of attack.

Table~\ref{tab:app_ood_full} presents the comprehensive evaluation of 12 models on AirfRANS extrapolation tasks. Beyond the Lift Coefficient ($C_L$) and its ranking correlation ($\rho_L$) reported in the main text, we additionally provide the Mean Squared Error (MSE) for surrounding (Volume) and surface (Surf) physics fields to offer a holistic view of model robustness.

\textbf{Performance Analysis.} 
Graph-based methods (e.g., MeshGraphNet~\cite{MeshGraphDBLP:conf/iclr/PfaffFSB21}, Graph U-Net~\cite{graphunet2019graph}) exhibit significant degradation in extrapolation settings. For instance, MeshGraphNet incurs a relative error of 1.7718 for $C_L$ in Reynolds extrapolation. Fundamentally, these Message Passing Neural Networks tend to overfit the local connectivity statistics of the training mesh. When the Reynolds number increases, the physical boundary layer thins, introducing sharp gradients that alter the effective physical length scales; Message Passing Neural Networks, constrained by learned local receptive fields, fail to adapt to this shift. While Transformer-based baselines like Transolver (++)~\cite{transolverDBLP:conf/icml/WuLW0L24,trans++DBLP:conf/icml/LuoWZXD0L25} and SAOT~\cite{SAOTzhou2025dual} improve upon this by leveraging global attention, they still rely heavily on latent feature similarity for token clustering. In scenarios with severe flow separation, the feature distribution shifts drastically, leading to unstable slicing and degraded field reconstruction.

In contrast, PGOT achieves strong stability across most metrics, obtaining the best performance in 7 out of 8 evaluation criteria. While Transolver (++) slightly outperforms PGOT on $C_L$ in the Unseen Angles setting (0.1503 vs. 0.1742), PGOT achieves superior ranking correlation ($\rho_L$: 0.9952 vs. 0.9948), which is more critical for design optimization as it measures the model's ability to correctly rank different airfoil designs. Moreover, PGOT consistently achieves the lowest field reconstruction errors (Volume and Surf) in both OOD settings, demonstrating that its geometric anchoring mechanism provides more reliable physical field predictions even when the coefficient prediction is slightly less accurate.

\section{Additional Visualizations}
\label{app:visualization}

\subsection{Learned Slice Assignments of SpecGeo-Attention}

\begin{figure*}[htbp]
    \centering
    \includegraphics[width=0.8\textwidth]{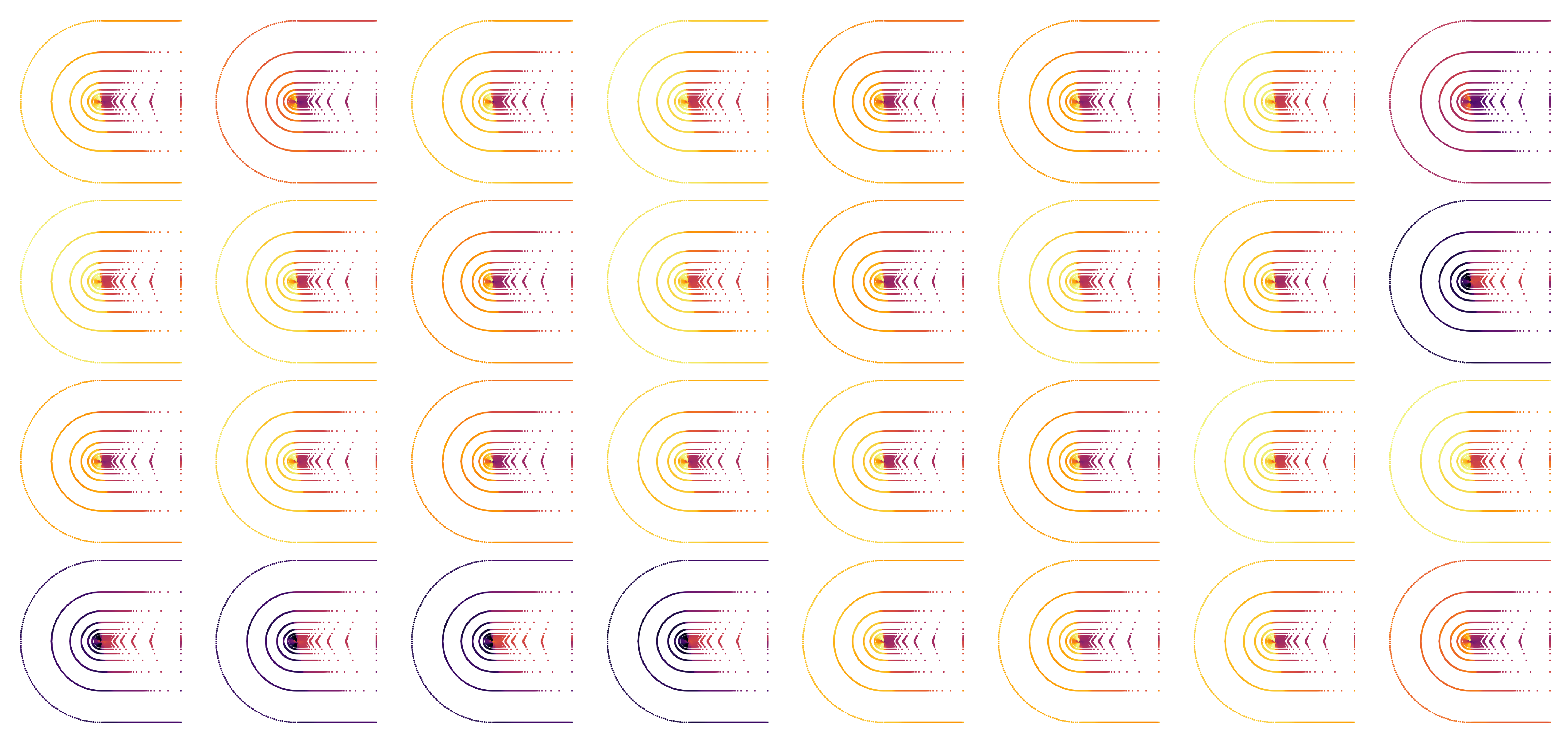}
    \caption{Learned slice assignments from SpecGeo-Attention on Airfoil (32 slices).}
    \label{fig:att_airfoil}
\end{figure*}

\begin{figure*}[htbp]
    \centering
    \includegraphics[width=0.8\textwidth]{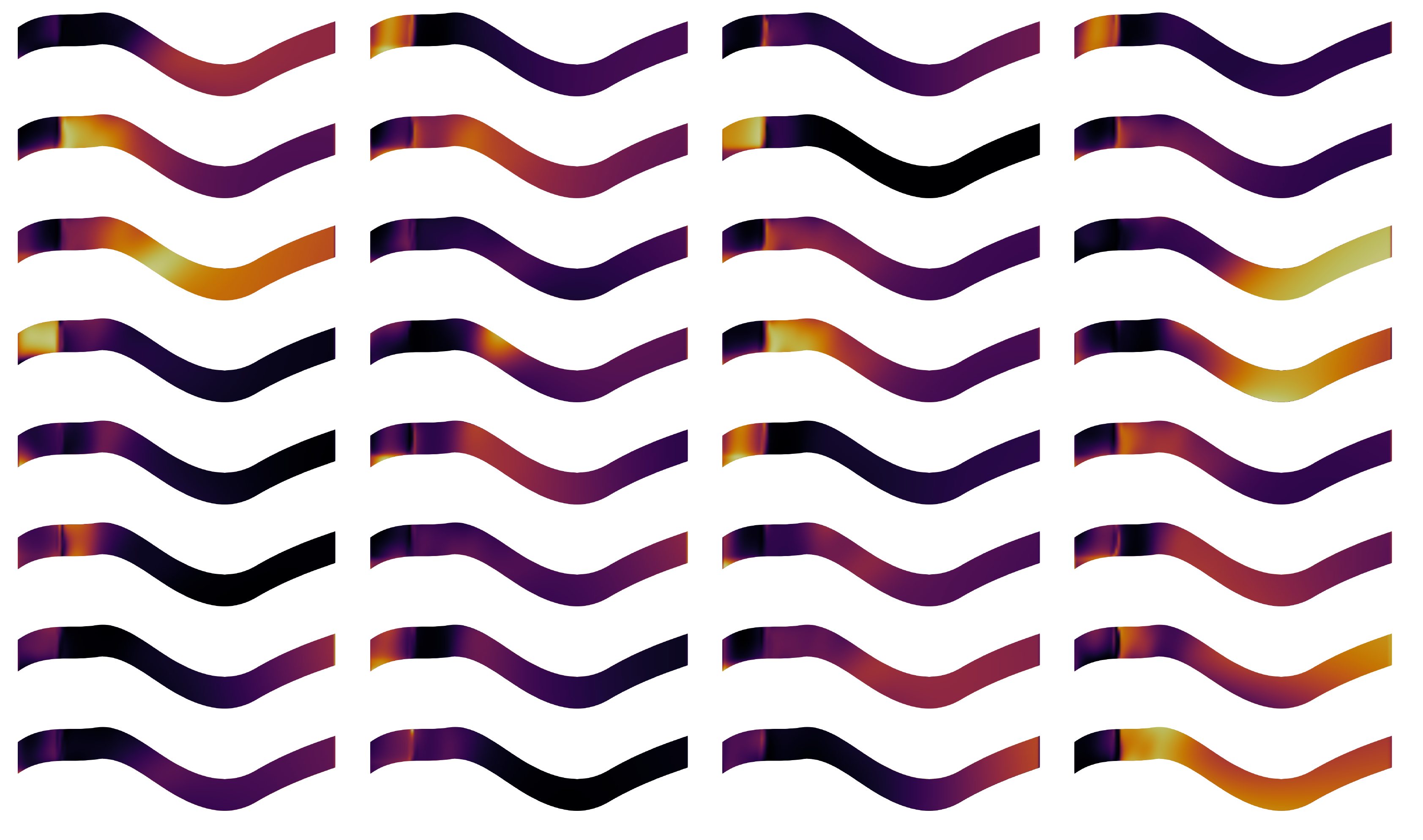}
    \caption{Learned slice assignments from SpecGeo-Attention on Pipe (32 slices).}
    \label{fig:att_pipe}
\end{figure*}

\begin{figure*}[htbp]
    \centering
    \includegraphics[width=\textwidth]{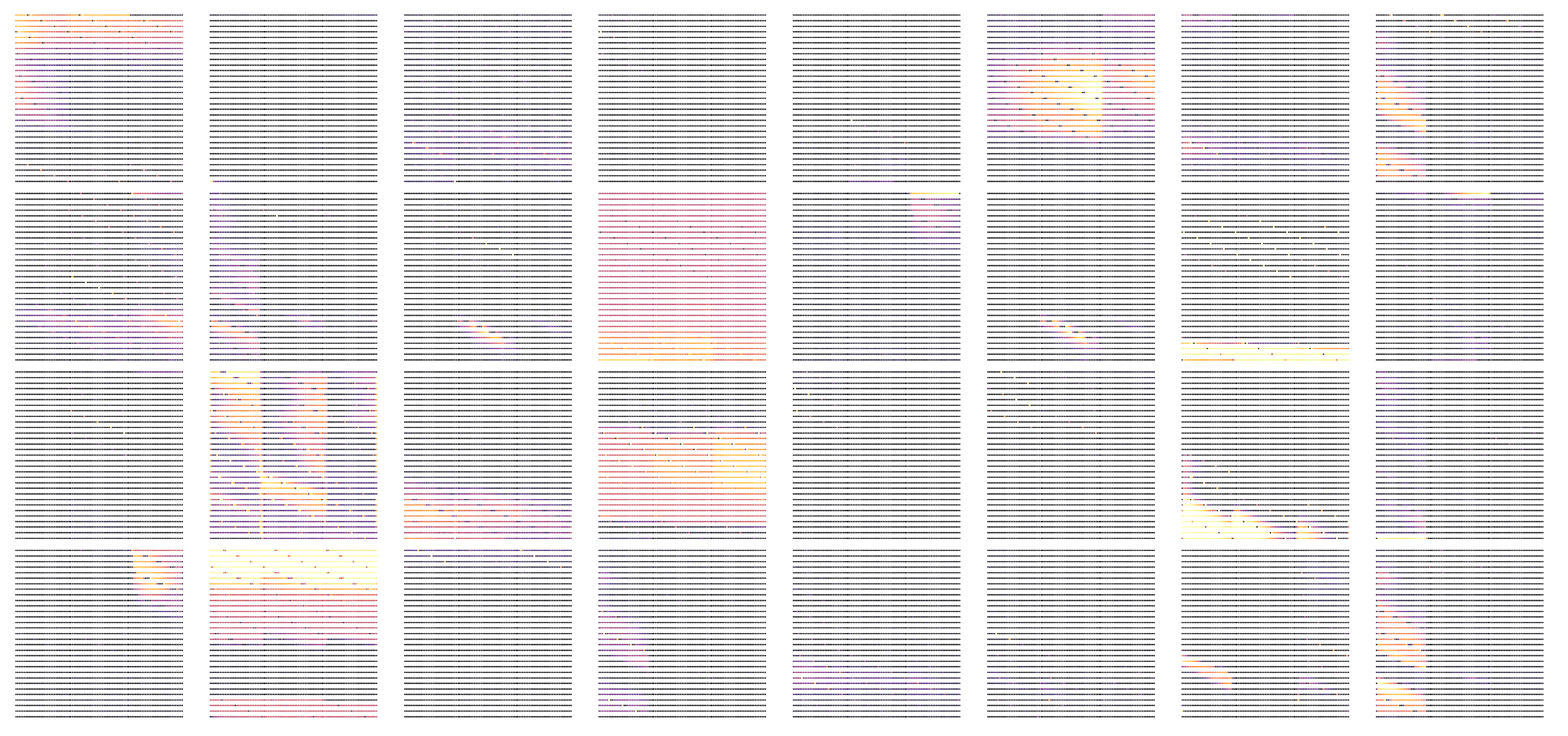}
    \caption{Learned slice assignments from SpecGeo-Attention on Plasticity (32 slices).}
    \label{fig:att_plasticity}
\end{figure*}

\begin{figure*}[htbp]
    \centering
    \includegraphics[width=\textwidth]{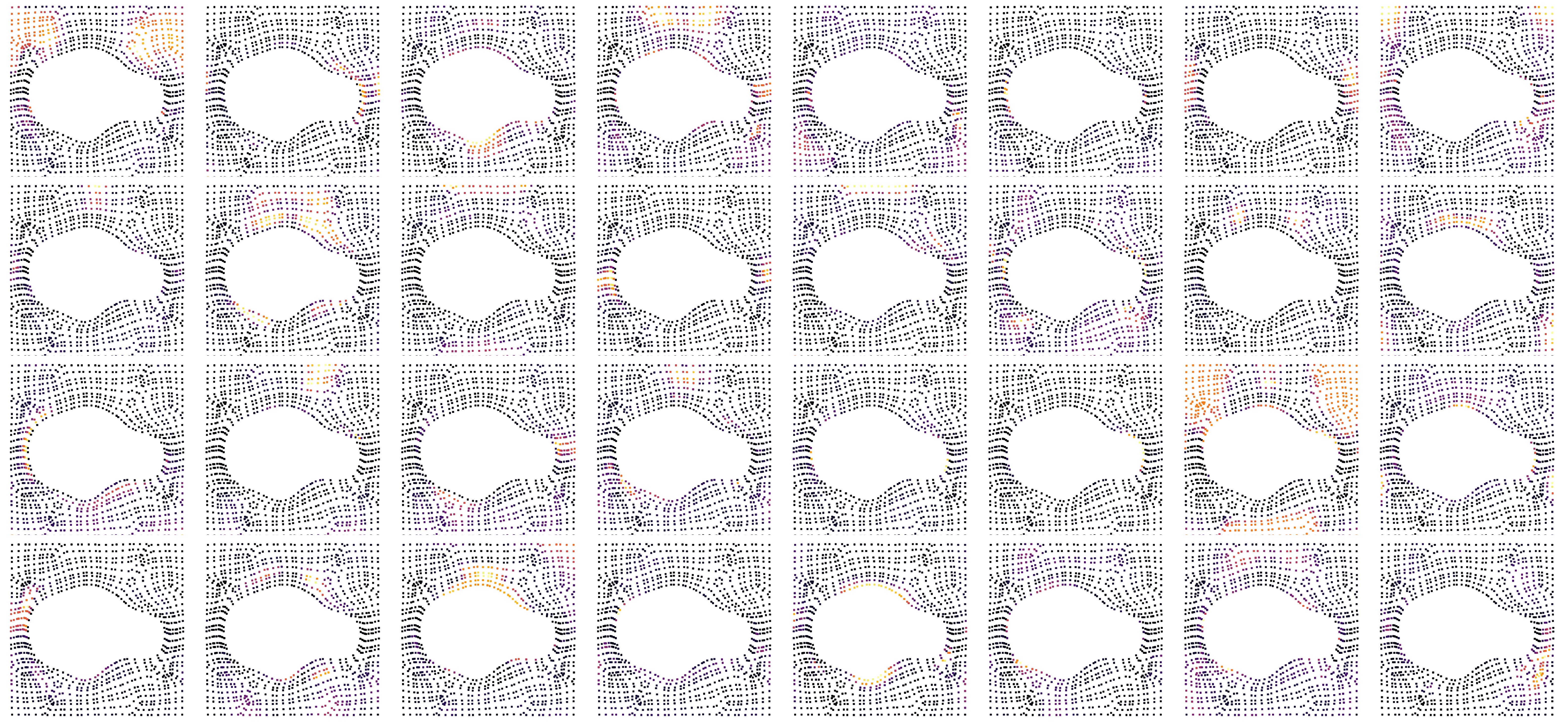}
    \caption{Learned slice assignments from SpecGeo-Attention on Elasticity (32 slices).}
    \label{fig:att_elasticity}
\end{figure*}

\begin{figure*}[htbp]
    \centering
    \includegraphics[width=0.9\textwidth]{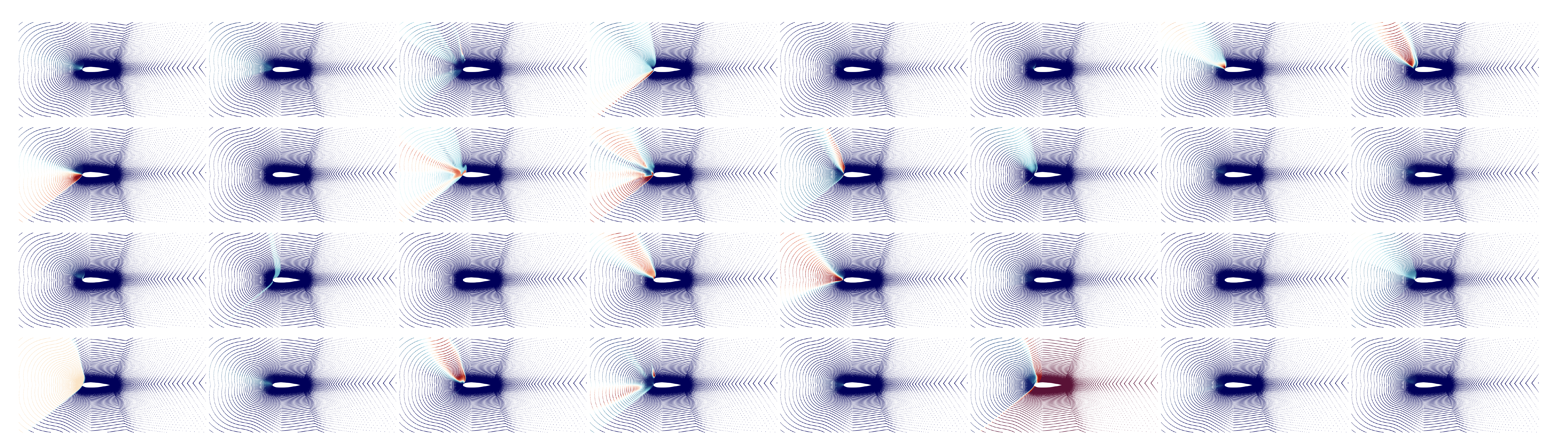}
    \caption{Learned slice assignments from SpecGeo-Attention on AirfRANS (32 slices).}
    \label{fig:att_airfrans}
\end{figure*}

\begin{figure*}[htbp]
    \centering
    \includegraphics[width=0.9\textwidth]{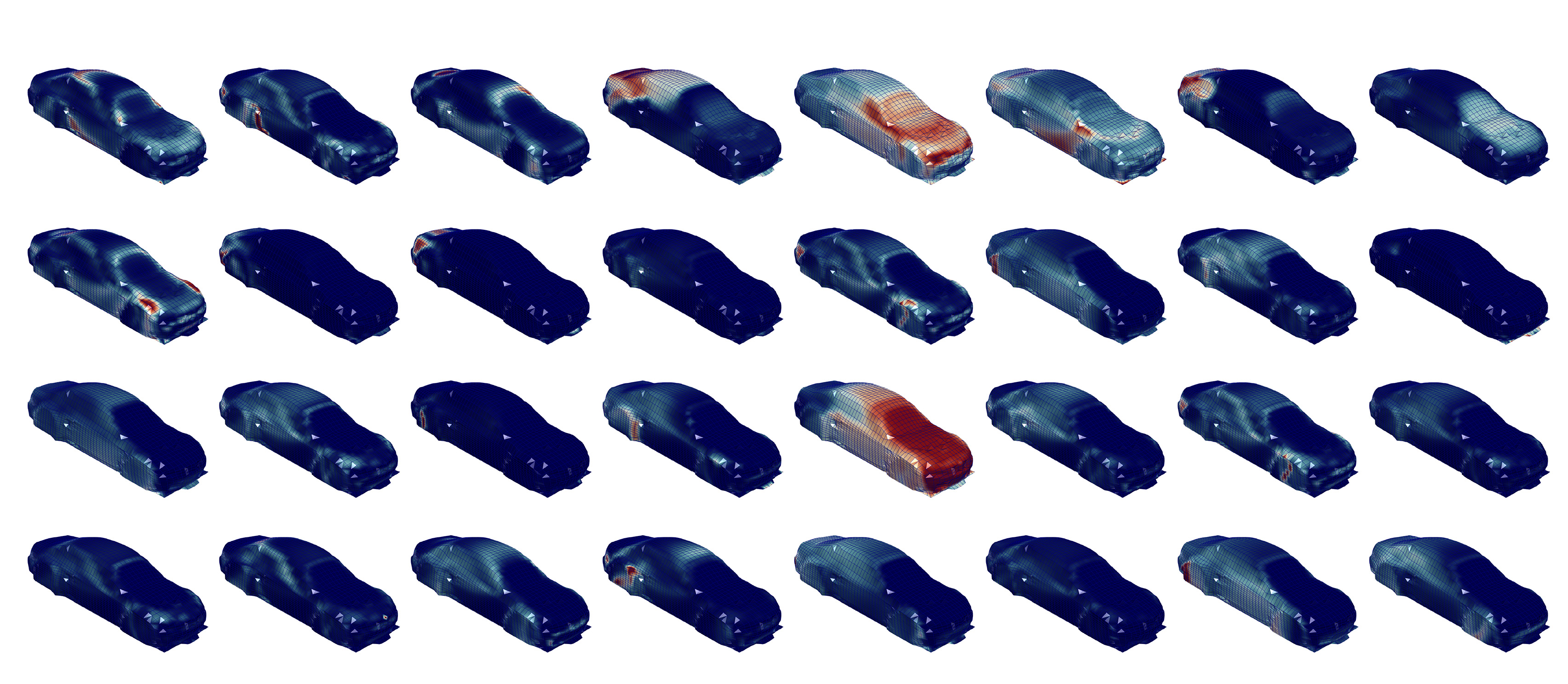}
    \caption{Learned slice assignments from SpecGeo-Attention on Shape-Net Car (32 slices).}
    \label{fig:att_car}
\end{figure*}

Figures~\ref{fig:att_airfoil}--\ref{fig:att_aircraft} visualize the learned slice assignments from SpecGeo-Attention across all seven benchmarks. Each subplot represents one of 32 latent physical states, where different colors indicate varying assignment intensities.
\subsection{Gate Activations of TaylorDecomp-FFN}

\begin{figure*}[htbp]
    \centering
    \includegraphics[width=0.9\textwidth]{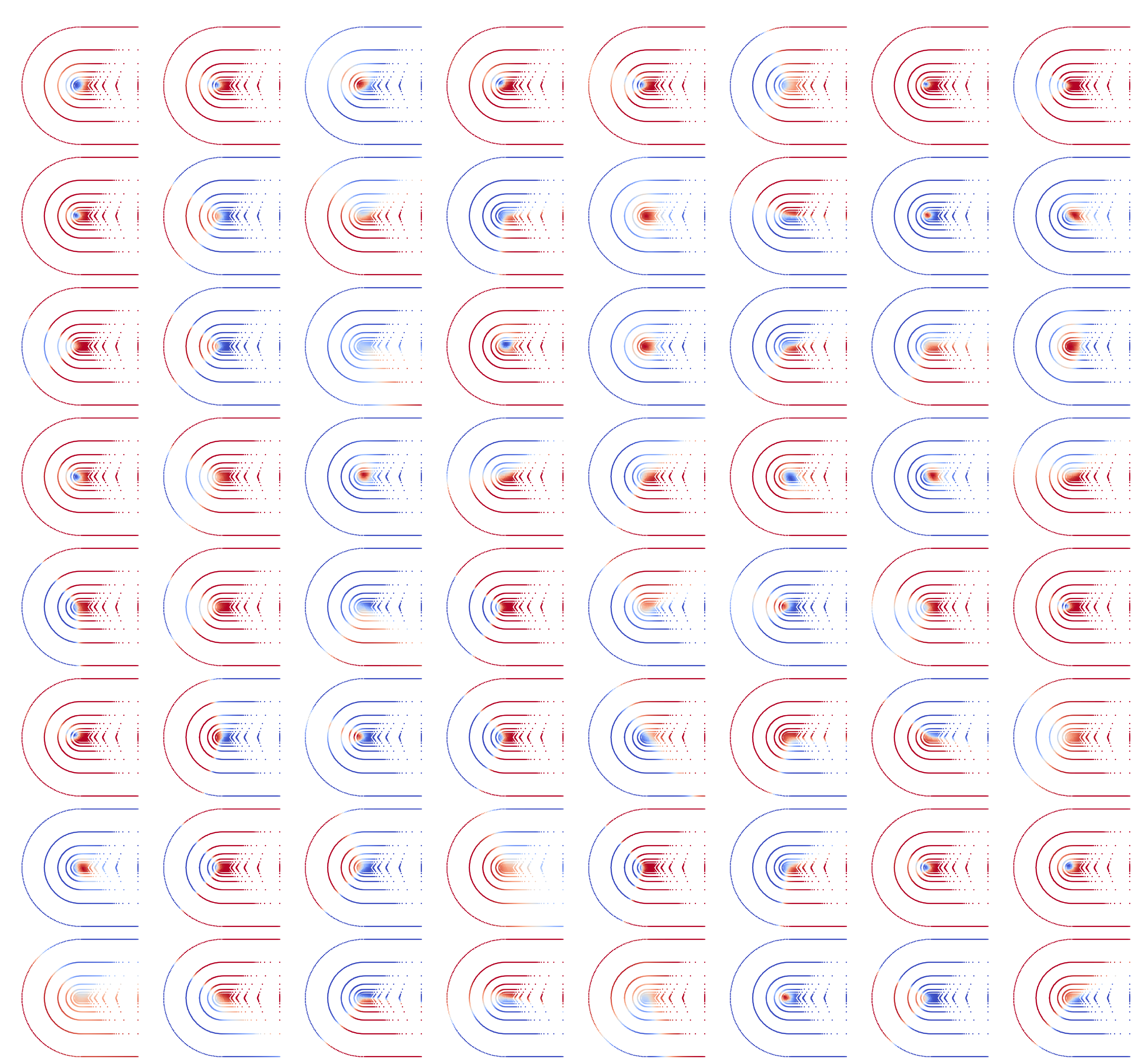}
    \caption{Gate activations from TaylorDecomp-FFN on Airfoil (64 channels).}
    \label{fig:gate_airfoil}
\end{figure*}

\begin{figure*}[htbp]
    \centering
    \includegraphics[width=0.9\textwidth]{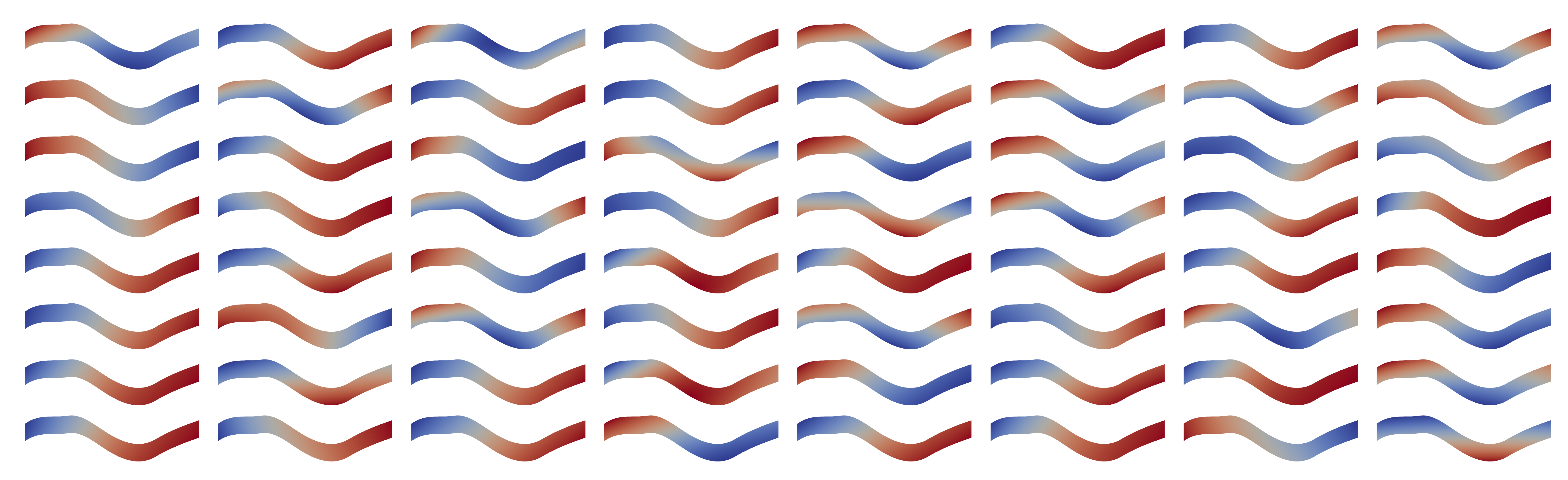}
    \caption{Gate activations from TaylorDecomp-FFN on Pipe (64 channels).}
    \label{fig:gate_pipe}
\end{figure*}

\begin{figure*}[htbp]
    \centering
    \includegraphics[width=0.9\textwidth]{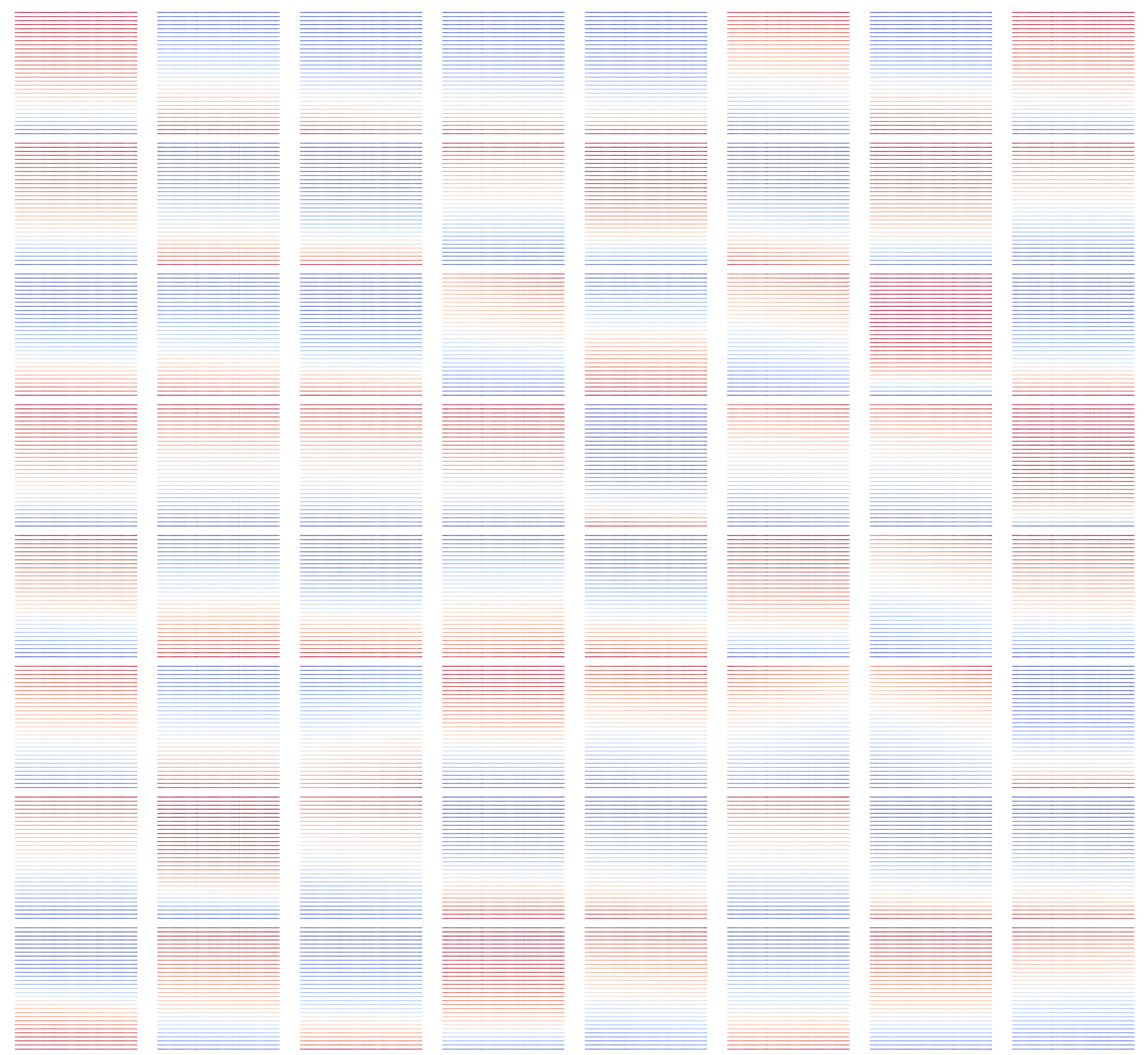}
    \caption{Gate activations from TaylorDecomp-FFN on Plasticity (64 channels).}
    \label{fig:gate_plasticity}
\end{figure*}

\begin{figure*}[htbp]
    \centering
    \includegraphics[width=0.9\textwidth]{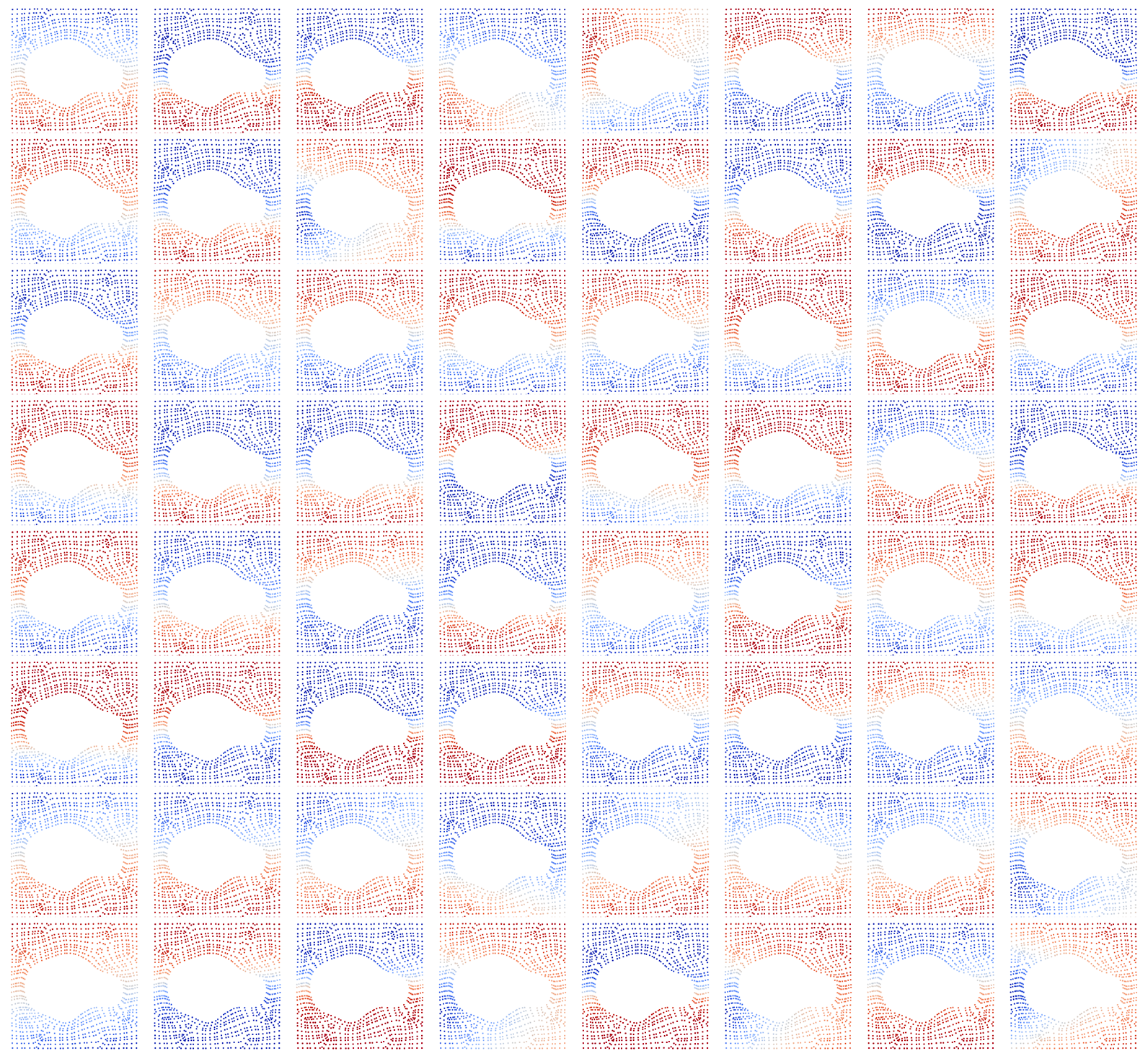}
    \caption{Gate activations from TaylorDecomp-FFN on Elasticity (64 channels).}
    \label{fig:gate_elasticity}
\end{figure*}

\begin{figure*}[htbp]
    \centering
    \includegraphics[width=0.9\textwidth]{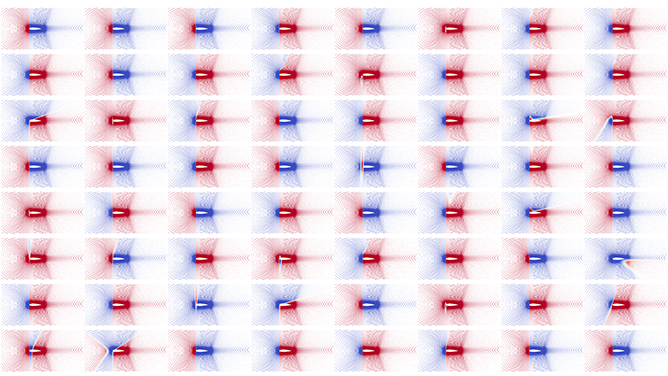}
    \caption{Gate activations from TaylorDecomp-FFN on AirfRANS (64 channels).}
    \label{fig:gate_airfrans}
\end{figure*}

\begin{figure*}[htbp]
    \centering
    \includegraphics[width=0.9\textwidth]{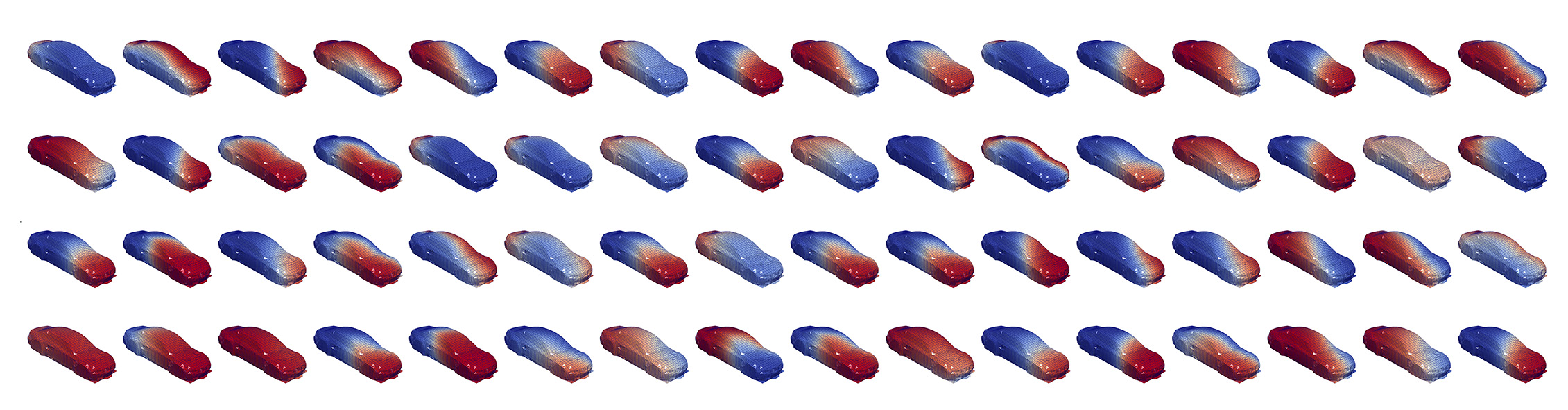}
    \caption{Gate activations from TaylorDecomp-FFN on Shape-Net Car (64 channels).}
    \label{fig:gate_car}
\end{figure*}

\begin{figure*}[htbp]
    \centering
    \includegraphics[width=\textwidth]{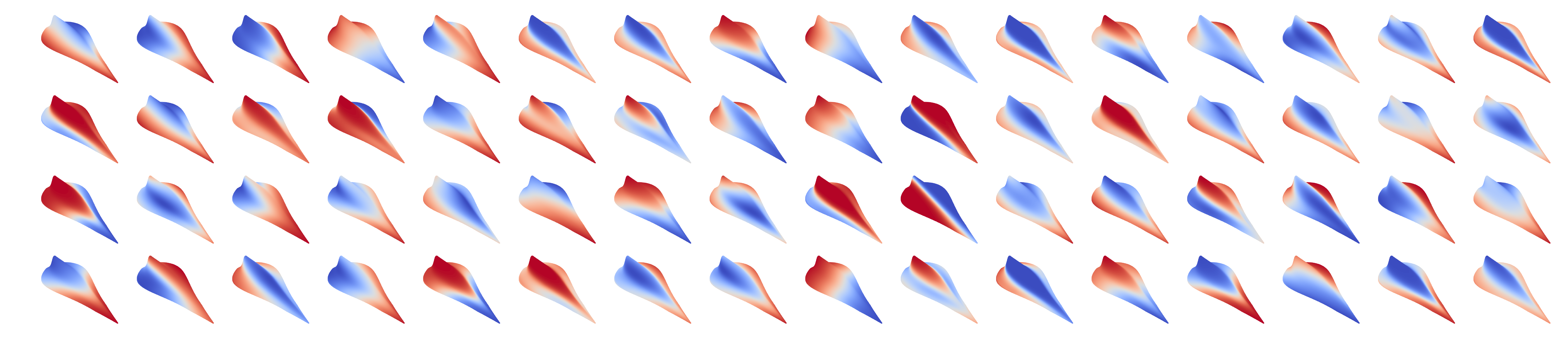}
    \caption{Gate activations from TaylorDecomp-FFN on AirCraft (64 channels).}
    \label{fig:gate_aircraft}
\end{figure*}

Figures~\ref{fig:gate_airfoil}--\ref{fig:gate_aircraft} visualize the gate activations $\boldsymbol{\alpha}(\mathbf{g})$ from TaylorDecomp-FFN. Each subplot displays the gate values for one of 64 feature channels. Red indicates high activation (routing to the non-linear expert) and blue indicates low activation (routing to the linear expert).
\clearpage

\begin{figure*}[htbp]
    \centering
    \includegraphics[width=0.85\textwidth]{att_AirCraft.png}
    \caption{Learned slice assignments from SpecGeo-Attention on AirCraft (32 slices).}
    \label{fig:att_aircraft}
\end{figure*}

\end{document}